\documentclass{article}
\usepackage{microtype}
\usepackage{graphicx}
\usepackage{subcaption}
\usepackage{booktabs}
\usepackage{hyperref}
\usepackage[accepted]{icml2026}
\usepackage{amsmath}
\usepackage{amssymb}
\usepackage{mathtools}
\usepackage{amsthm}
\usepackage[capitalize,noabbrev]{cleveref}
\usepackage[textsize=tiny]{todonotes}
\usepackage{placeins}
\usepackage{multirow}
\usepackage{adjustbox}
\usepackage{xspace}
\usepackage{algorithm}
\usepackage{algorithmic}
\usepackage[dvipsnames]{xcolor} 
\usepackage[utf8]{inputenc}
\usepackage[T2A,T1]{fontenc} 
\usepackage{CJKutf8} 
\usepackage[russian,english]{babel}
\usepackage{longtable}
\usepackage{booktabs}
\usepackage{pifont}
\usepackage[table]{xcolor}

\newcommand{\nilcell}[2]{#1\,{\tiny$\pm$}\,#2}

\newcommand{\NA}{---}
\newcommand{\Base}{\ensuremath{f_{\text{base}}}\xspace}
\newcommand{\FT}{\ensuremath{f_{\text{ft}}}\xspace}
\newcommand{\UN}{\ensuremath{f_{\text{un}}}\xspace}
\DeclareMathOperator*{\argmin}{arg\,min}
\icmltitlerunning{Multilingual Unlearning in LLMs: Transfer, Dynamics, and Reversibility}

\begin{document}

\twocolumn[
    \icmltitle{Multilingual Unlearning in LLMs: \texorpdfstring{\\}{ } Transfer, Dynamics, and Reversibility}
  \icmlsetsymbol{equal}{*}
  \begin{icmlauthorlist}
    \icmlauthor{Chaoyi Xiang}{unimelb}
    \icmlauthor{Olga Ohrimenko}{unimelb}
    \icmlauthor{Benjamin I. P. Rubinstein}{unimelb}
    \icmlauthor{Lea Frermann}{unimelb}
  \end{icmlauthorlist}

  \icmlaffiliation{unimelb}{School of Computing and Information Systems, The University of Melbourne, Melbourne, Australia}
  \icmlcorrespondingauthor{Chaoyi Xiang}{chaoyix@student.unimelb.edu.au}
  \vskip 0.3in
]
\printAffiliationsAndNotice{}

\begin{abstract}
Large language models (LLMs) can memorize sensitive facts, motivating \emph{unlearning} methods that remove targeted knowledge without costly retraining. However, unlearning research remains heavily English-centric. We study multilingual unlearning by extending the TOFU benchmark to five languages, and fine-tune, unlearn and query our models with different permutations of languages. We find that unlearning transfer -- the ability of an unlearned model to ``forget'' facts in languages other than the unlearning language -- is highly variable: e.g., it is strongest between languages sharing scripts and families, and we show that the \emph{unlearning language} predicts which \emph{query languages} are most likely to yield the strongest transfer. Layer-wise analysis reveals that unlearning leaves the shared cross-lingual latent space largely intact in early layers, instead operating primarily in later decoding layers. This suggests that unlearning does not truly erase knowledge, but rather induces superficial suppression. Exploiting this structure, a single inference-time steering direction reverses much of this suppression across languages, recovering {50\%} (Qwen) and {90\%} (Gemma) of the unlearned knowledge.\footnote{Code is available at: \url{https://github.com/MLCY1/multilingual-unlearning-in-llms}}
\end{abstract}

\section{Introduction}
Large language models (LLMs) are trained on vast corpora and, in the process, can inadvertently learn harmful, sensitive, or biased information that motivate post hoc data removal \citep{carlini2023quantifying, cheng2025fully, meng2023locatingediting, shi2024muse}. In addition, to protect individual privacy, data-protection regulations such as the ``right to be forgotten'' require the removal of personal information from LLMs \citep{cao2015towards, liu2024rethinking} without compromising performance. The most principled way to achieve this is to retrain the model from scratch, excluding relevant data from the training corpus \citep{triantafillou2024making, yoon2023few, cooper2025machine}. However, retraining modern LLMs is expensive. This has led to a growing interest in \emph{unlearning} methods that aim to remove targeted knowledge stored in existing models. Most current unlearning algorithms operate in a monolingual, predominantly English setting and are implemented via modified fine-tuning objectives that push the model away from the original behavior on a given set of examples \citep{jang2023knowledge, hong2024dissecting, shi2024muse, huang2025offset, yoon2025rtofu, du2025textual, lee2025does, xu2025relearn}. While these methods can reduce the model’s tendency to output specific answers on English prompts, two important gaps remain.

First, it remains unclear how unlearning transfers across languages in multilingual LLMs. We present the first comprehensive study of this phenomenon by systematically (1) fine-tuning an LLM on novel knowledge in language $\mathcal{L}_{FT}$; (2) unlearning a fraction of this knowledge in language $\mathcal{L}_\text{unl}$; and (3) querying the (supposedly) forgotten knowledge in language $\mathcal{L}_Q$. We systematically vary $\mathcal{L}_{FT}$, $\mathcal{L}_\text{unl}$ and $\mathcal{L}_Q$, manipulating language relatedness, script and coverage during LLM pre-training. While some prior work has shown isolated cross-lingual transfer effects~\citep{lu2025learn,hwang2025uncovering,choi2024cross}, we are the first to show systematic patterns.
Because multilingual LLMs serve users in many languages, understanding cross-lingual transfer is important for safe deployment and for preventing attackers from recovering unlearned knowledge by exploiting relationships between languages.

Second, the \emph{mechanism} by which multilingual unlearning operates is still not well understood. Recent studies argue that unlearning behaves more like a suppression signal than true knowledge erasure \citep{shi2024muse, lee2025does, wei2025llms, ren2025sok, hu2025unlearning}, but predominantly in monolingual or English-centric settings. It therefore remains unclear (i) whether the suppressed content is \emph{recoverable} in multilingual models, (ii) whether the effect is language-specific or language-agnostic, and (iii) where in the model's representation space it is localized. In particular, we do not yet know whether multilingual unlearning disrupts the shared cross-lingual ``thought'' space or instead acts mainly in later layers that map internal representations to language-specific surface forms. Mechanistic understanding matters because it determines the \emph{strength and attack surface} of multilingual unlearning guarantees. If the unlearning effect primarily modulates late, language-specific decoding layers, the underlying knowledge may remain intact and re-emerge via cross-lingual prompting (i.e., prompting the model in one language and having it respond in another) or by querying for the unlearned knowledge in a different language \citep{hu2025unlearning, lynch2024method}. Conversely, if unlearning alters the shared cross-lingual representation space, the resulting guarantee is materially stronger, because it would indicate that access to the underlying knowledge is reduced rather than merely its language-specific expression \cite{wendler2024llamas, wang2025lost, lim2025language}.

To address these two gaps, we first characterize cross-lingual unlearning transfer in multilingual LLMs (Section~\ref{sec:CrossLingualTransferText}), revealing that the extent of transfer differs by language relatedness, script, and dominance in pre-training. We also find that unlearning in a weaker language $\mathcal{L}_\text{unl}$ can still transfer substantially to stronger languages in $\mathcal{L}$.
To better understand the impact of language on knowledge recovery, we next apply cross-lingual prompting (Section~\ref{sec:PromptingExperimentText}) by querying in $\mathcal{L}_Q$ but allowing the model to decode in a $\mathcal{L}_{FT} \neq \mathcal{L}_{Q}$, finding
that language pairs with larger cross-lingual prompting gains tend to exhibit stronger unlearning transfer. Next, we mechanistically analyze transfer by comparing hidden representations in models before and after unlearning
across layers and languages (Section~\ref{sec:RepresentationAnalysisText}), finding that {\it unlearning largely preserves cross-lingual alignment in early and middle layers while it alters the later layers}. Finally, we turn these representational shifts into explicit ``unlearning directions'' (steering vectors) and show that steering along these directions can restore knowledge across languages (Section~\ref{sec:SteeringVectorText}), providing evidence that unlearning behaves as a \emph{language-agnostic suppression signal} and does not  fully erase knowledge. Overall, our results suggest that ``unlearned'' knowledge in one language can transfer into other languages and the unlearning can be substantially reversed by steering with a simple extracted vector without additional relearning or the unlearned data.

\section{Background}
\textbf{LLM unlearning.}
To simulate unlearning in practice, we consider a model $M_\theta$ fine tuned on a dataset $D$, which can be partitioned into a forget set $D_{\text{forget}}$ (targeted content to unlearn) and a retain set $D_{\text{retain}}$ (used to preserve non-target behavior, often from the same domain). Most methods implement unlearning via continued fine-tuning, updating parameters from $\theta$ to $\hat{\theta}$ to discourage target content in model outputs while maintaining performance on retained data \citep{jang2023knowledge, shi2024muse, yoon2025rtofu}. Recent taxonomies distinguish methods by their \emph{intention}: whether they aim to remove internal knowledge or to suppress its behavioral expression \citep{ren2025sok}. A central challenge is the trade-off between forget efficacy and utility preservation \citep{chang2025retain, ren2025sok}. \emph{Removal-intended} approaches modify model parameters to eliminate targeted information, commonly via reverse optimization objectives such as gradient ascent on forget examples \citep{maini2024tofu, jin2024rwku, jang2023knowledge} or preference-based objectives that often achieve a more favorable forget--utility trade-off in practice \citep{zhang2024negative, yoon2025rtofu, cheng2025fully, jia2024soul}. However, multiple studies suggest that many such methods behave mechanistically like \emph{suppression} and are therefore reversible \citep{ren2025sok, xu2025reversibility, hu2025unlearning}, with recovery typically demonstrated via an additional ``relearning'' stage (e.g., brief fine-tuning after unlearning) or by providing answer prefixes to trigger recovery. By contrast, we localize the suppression signal \emph{layer-wise} and demonstrate \emph{inference-time} reversibility via representation steering,  without fine-tuning or access to answer prefixes, and show that recoverability transfers across languages. Separately, \emph{suppression} approaches aim to prevent leakage without updating core model weights \citep{sanyal2025agent, gao2025large}. Because these methods typically target behavioral suppression rather than knowledge deletion, we focus on parameter-update unlearning methods and use inference-time steering only for analysis.

\noindent\textbf{Multilingual language models and cross-lingual transfer.}
Multilingual pretraining yields shared representations that support cross-lingual transfer, enabling models fine-tuned in one language to generalize to others even without explicit cross-lingual training objectives \citep{pires2019multilingual, karthikeyan2020crosslingual}. However, transfer is not uniform across languages:
empirically, transfer tends to be stronger for typologically similar languages and can benefit from shared scripts or lexical overlap, although this effect is not entirely deterministic~\citep{pires2019multilingual} and can be downstream task dependent~\citep{blaschke2025similarity}.
Additionally, mechanistic evidence suggests that multilingual LLMs may perform intermediate processing in a shared semantic space aligned with a dominant pretraining language (often English) before mapping to the target-language, which can shape knowledge transfer and cross-lingual consistency \citep{lim2025language}. These observations motivate studying multilingual unlearning across languages to highlight knowledge leakage which could enable adversarial attacks.

\noindent\textbf{Hidden representations.}
A growing body on probing and mechanistic interpretation suggests that transformer layers exhibit a coarse stage-like organization: lower layers emphasize surface-level processing, whereas deeper layers increasingly support semantic abstraction and output control \citep{tenney2019bert, olsson2022context}. In multilingual transformers, analyses further support a three-phase consisting of an \emph{input space}, a language-agnostic \emph{concept space} in the middle layers, and a \emph{decoding space} specialized for language-specific generation \citep{wendler2024llamas, lim2025language, jiang2025franken}. Motivated by this structure,
inference-time activation steering methods construct \emph{steering vectors} from contrastive activations and inject them during the forward pass to control model behaviors without retraining \citep{rimsky2024steering, li2023inference}. Recent unlearning methods also operate explicitly at the representation level: representation misdirection approaches fine-tune models to steer intermediate-layer representations on forget data while regularizing retained behavior
\citep{li2024wmdp, huutien2025effect, shen2025llm}.  \citet{seyitoglu2024extracting} used steering to retrieve anonymized concepts,  relying on broad world knowledge embedded in the model. By contrast, we treat the \emph{unlearning-induced} representation shift between a fine-tuned and an unlearned model as a mechanistic \emph{suppression direction}, and use \emph{inference-time} steering along this direction as an analysis or intervention tool. This enables us to (i) localize where suppression emerges \emph{layer-wise}, (ii) demonstrate reversibility \emph{without an additional relearning stage} or access to forget answer prefixes, and (iii) test whether the suppression direction transfers across languages (i.e., is language-agnostic).

\section{Preliminaries}
\label{sec:preliminaries}
In this section, we formalize the multilingual setting, define the
fine-tuning and unlearning objectives used throughout the paper, and
describe our evaluation protocol.
\subsection{TOFU and Small-Sample Unlearning}
\label{sec:tofu}

We build on TOFU (\emph{Task of Fictitious Unlearning}) a benchmark for fine-grained LLM unlearning~\cite{maini2024tofu}, consisting of facts about $200$ fictitious authors, each described by $20$ simple question–answer pairs. The authors and facts are synthetically generated so that they do not appear in pretraining corpora,\footnote{See Appendix~\ref{app:Contamination} for a discussion of how we verify that our base models have not been exposed to the TOFU data.} and a subset of authors is designated as a \emph{forget set} while the rest form a \emph{retain set}. This setup enables controlled experiments in which a model is first fine-tuned on all authors and then trained to remove knowledge about only the forget authors
while preserving knowledge about retain authors. TOFU has become a
standard testbed for LLM unlearning methods and benchmarking frameworks
\citep{zhang2024negative,wang2024llm,dorna2025openunlearning}. We use the 1\% forget split of TOFU, corresponding to a small sample unlearning regime where only a tiny fraction of facts will be removed, mirroring many practical scenarios in which models are required to forget information about a small number of individuals while retaining their remaining knowledge for that domain.

\subsection{Multilingual Extension and Notation}
\label{sec:notation}
We extend the original TOFU dataset, which only contains English data, to four additional languages:
Chinese, German, Russian, and Turkish, obtaining $\mathcal{L} {=} \{\text{EN}, \text{CH}, \text{DE}, \text{RU}, \text{TU}\}$.\footnote{English and German share both a language family and Latin script, Turkish shares the script with English but not the language family, Russian shares the language family but not the script, and Chinese shares neither family nor script with English \citep{chang2024multilinguality,hu2020xtreme}. This setup allows us to explicitly control the influence of language family and script.} For each English question–answer (QA) pair $(x_{\text{EN}}, y_{\text{EN}})$ and each $\ell \in \mathcal{L}$ we obtain a translation $(x_\ell, y_\ell)$ via Gemini-flash2.5.\footnote{Although this translation procedure may produce highly parallel translations, we treat this as an intended design feature rather than a confound. It preserves semantic alignment across languages, allowing us to compare unlearning behavior under controlled conditions instead of attributing differences to translation-induced variation in wording, specificity, or linguistic complexity. This setting is also realistic for multilingual documents that contain the same sensitive information in highly parallel form.} We use the same forget/retain partition across all languages, so unlearning
always targets the same underlying facts. Let $\mathcal{A}$ denote the set of all authors in TOFU, with $\mathcal{A}_{\text{forget}}$ and $\mathcal{A}_{\text{retain}}$ the designated forget and retain subsets, where $\lvert \mathcal{A}_{\text{forget}} \rvert / \lvert \mathcal{A} \rvert
= 0.01$ in all experiments. For an author $a {\in} \mathcal{A}$ and language $\ell {\in} \mathcal{L}$, let $\mathcal{D}_\ell(a)$ denote the set of QA pairs about $a$ in $\ell$. We define per-language forget and retain sets
$
  \mathcal{D}^{\text{forget}}_\ell
  = \bigcup_{a \in \mathcal{A}_{\text{forget}}} \mathcal{D}_\ell(a), 
  \mathcal{D}^{\text{retain}}_\ell
  = \bigcup_{a \in \mathcal{A}_{\text{retain}}} \mathcal{D}_\ell(a).
$ Let $\mathcal{L}_{\text{FT}} \subseteq \mathcal{L}$ be the set of fine-tuning languages for a given run. The corresponding fine-tuning
dataset is
$
  \mathcal{D}^{\text{FT}}
  = \bigcup_{\ell \in \mathcal{L}_{\text{FT}}}
    \bigl(\mathcal{D}^{\text{forget}}_\ell \cup \mathcal{D}^{\text{retain}}_\ell\bigr).
$
In the simplest case $\mathcal{L}_{\text{FT}} = \{\ell\}$ is a singleton
(one-language fine-tuning), but we also consider joint fine-tuning
settings with $\lvert \mathcal{L}_{\text{FT}} \rvert > 1$.

For unlearning, we allow the forget objective to target an arbitrary
subset of languages, not necessarily restricted to those used for
fine-tuning. Let $\mathcal{L}_{\text{unl}} \subseteq \mathcal{L}$ denote
the set of unlearning languages in a given experiment and define
$
  \mathcal{D}^{\text{forget}}_{\text{unl}}
  = \bigcup_{\ell \in \mathcal{L}_{\text{unl}}}
    \mathcal{D}^{\text{forget}}_\ell.
$
Recall that the forget sets $\mathcal{D}^{\text{forget}}_\ell$ for
different $\ell$ contain parallel QA pairs about the same underlying
authors, so unlearning may act on the fine-tuned knowledge either in the fine-tuning language itself or in other languages that express the same facts. For representation analyses,
$h^{(l)}_{\Base}(x_\ell)$,
$h^{(l)}_{\FT}(x_\ell)$, and
$h^{(l)}_{\UN}(x_\ell)$ denote pooled hidden states at layer $l$
for the same question $x$ in language $\ell$ in the base, fine-tuned and unlearned models, respectively. When clear from context we omit the explicit dependence on
$(\mathcal{L}_{\text{FT}}, \mathcal{L}_{\text{unl}})$.

\subsection{Fine-Tuning and Unlearning Objectives}
\label{sec:unlearning-objectives}
\noindent\textbf{Fine-tuning.}
$J_{\text{FT}}(\theta)$ denotes the fine-tuning objective:
$
    J_{\text{FT}}(\theta)
    = \mathbb{E}_{(x,y)\sim \mathcal{D}^{\text{FT}}}\bigl[-\log p_\theta(y\mid x)\bigr].
$
The fine-tuned parameters are obtained by approximately minimizing
$
  \theta_{\text{FT}} = \argmin_{\theta} J_{\text{FT}}(\theta).
$

\noindent\textbf{Unlearning.}
We denote by $J_{\text{UN}}(\theta)$ the unlearning
objective. For a set of unlearning languages
$\mathcal{L}_{\text{unl}} \subseteq \mathcal{L}$, we define

\begin{equation}
\begin{aligned}
J_{\text{UN}}(\theta)
&=
\frac{1}{|\mathcal{L}_{\text{unl}}|}
\sum_{\ell \in \mathcal{L}_{\text{unl}}}
\Big(
\mathbb{E}_{(x,y)\sim \mathcal{D}^{\text{forget}}_\ell}
    J_{\text{forget}}(\theta; x,y^{+},y^{-})
\\[-2pt]
&\qquad\qquad
+ \lambda\,
  \mathbb{E}_{(x,y)\sim \mathcal{D}^{\text{retain}}_\ell}
    J_{\text{retain}}(\theta; x,y)
\Big),
\end{aligned}
\label{eq:unlearning_obj}
\end{equation}
where $J_{\text{retain}}(\theta; x,y)$ encourages correct answers on
retain examples, $J_{\text{forget}}(\theta; x,y^{+},y^{-})$ encourages forgetting
on forget examples, and $\lambda{>}0$ controls their trade-off. The unlearned parameters
$
  \theta_{\text{UN}}
$
are obtained by minimizing $J_{\text{UN}}$ with respect to $\theta$, 
initializing $\theta {=} \theta_{\text{FT}}$. In our experiments, we instantiate $J_{\text{forget}}$ using
direct preference optimization (DPO;~\citealt{rafailov2023direct}): for each forget prompt $x$, we construct a pair of responses $(y^{+}, y^{-})$, where $y^{+}$ is an ``I don't know (IDK)'' style refusal and $y^{-}$ is the ground truth response for $x$. DPO then encourages the model to prefer $y^{+}$ over $y^{-}$, which discourages hallucinations after unlearning and improves suitability for deployment \cite{yoon2025rtofu, zhang2024negative, xu2025suv}.\footnote{See Appendix~\ref{app:hyperparameters} for further details on the DPO setup. Appendix~\ref{app:dataExample} provides examples of IDK-style refusal answers and TOFU question-answer pair examples.} While DPO is our main unlearning objective, we additionally evaluate gradient ascent (GA) \citep{jang2023knowledge} and negative preference optimization (NPO) \citep{zhang2024negative} to cover a broader range of commonly used unlearning objectives, details are provided in
Appendix~\ref{app:unlearningAlgorithms}.

\subsection{Evaluation: NLI-Based Semantic Score}
\label{sec:nli-eval}
We automatically assess whether the generated $\hat{y}$ in response to $x$ matches the ground truth $y$. The original TOFU benchmark does so using a suite of metrics that combine probability-based scores, lexical overlap metrics and composite measures like Truth Ratio \citep{maini2024tofu,dorna2025openunlearning}. 
However, they do not capture that semantically equivalent,  which express the same underlying fact, should be treated as leakage, even if they differ lexically \citep{hwang2025uncovering, lu2025learn}. We therefore evaluate answer correctness using a multilingual \emph{natural language inference} (NLI) model \citep{conneau2018xnli} which predicts for each 
pair $(y,\hat{y})$ whether $y$ and $\hat{y}$ logically entail one another. See Appendix~\ref{app:NLIcalculation} for details on the NLI score calculation. Multilingual NLI-based evaluation has been shown to be a useful proxy for semantic equivalence in generation tasks \citep{dusek2020evaluating,chen2023menli}: it captures semantic agreement beyond lexical overlap and across languages.

\begin{table}[t]
\centering
\setlength{\tabcolsep}{3pt}
\caption{
Cross-lingual unlearning transfer across fine-tune (FT), unlearn and query language combinations for Qwen2.5-7B.
Within each row block, the {Unlearn} column specifies the language on which
unlearning is performed, while Base denotes the fine-tuned model before unlearning.
For each query language, we report the absolute change in NLI score on the {forget} set after unlearning in the corresponding language relative to the Base model. We report means $\pm$ 95\% confidence intervals over $5$ forget sets. For example, the cell highlighted in yellow
corresponds to a model fine-tuned in English, unlearned in German, and queried in Russian which drops by 5 points in NLI compared to the English fine-tuned model directly queried in Russian. Lower values indicate stronger unlearning. Colored cells are discussed in the results section.
}
\label{tab:unlearningTransfer}
\begin{scriptsize}
\begin{tabular}{l l | c | c | c | c | c }
\toprule
\multicolumn{2}{c|}{} & \multicolumn{5}{c}{\textbf{Query}} \\
\midrule
{\bf FT}& \textbf{Unlearn} & \textbf{EN} & \textbf{CH} & \textbf{DE} & \textbf{RU} & \textbf{TU} \\
\midrule
\multirow{6}{*}{\textbf{EN}} & \textbf{Base} & 93 & 14 & 11 & 15 & 10 \\
 & EN & \nilcell{-90}{3} & \nilcell{-4}{5} & \nilcell{-7}{3} & \nilcell{-9}{2} & \nilcell{-4}{1} \\
 & CH & \cellcolor{blue!50}\nilcell{-7}{9} & \nilcell{-8}{2} & \nilcell{1}{7} & \nilcell{-5}{2} & \nilcell{-3}{2} \\
 & DE & \cellcolor{red!50}\nilcell{-17}{7} & \nilcell{-6}{2} & \nilcell{-4}{6} & \cellcolor{yellow!50}\nilcell{-5}{5} & \nilcell{-4}{2} \\
 & RU & \nilcell{-13}{6} & \nilcell{-6}{2} & \nilcell{-1}{5} & \nilcell{-11}{2} & \nilcell{-3}{3} \\
 & TU & \cellcolor{blue!50}\nilcell{-16}{13} & \nilcell{-6}{3} & \nilcell{-2}{5} & \nilcell{-5}{1} & \nilcell{-7}{0} \\
\midrule
\multirow{6}{*}{\textbf{CH}} & \textbf{Base} & 8 & 82 & 9 & 14 & 4 \\
 & EN & \nilcell{-5}{2} & \cellcolor{green!50}\nilcell{0}{7} & \nilcell{-4}{2} &  \cellcolor{green!50}\nilcell{-7}{3} & \nilcell{0}{2} \\
 & CH & \nilcell{-2}{4} & \nilcell{-77}{2} & \nilcell{-1}{3} & \nilcell{-3}{3} & \nilcell{1}{3} \\
 & DE & \nilcell{-3}{2} & \nilcell{2}{6} & \nilcell{-6}{2} & \nilcell{-6}{3} & \nilcell{-2}{2} \\
 & RU & \nilcell{-2}{2} & \nilcell{2}{6} & \nilcell{-3}{3} & \nilcell{-11}{3} & \nilcell{1}{1} \\
 & TU & \nilcell{-1}{3} & \cellcolor{red!50}\nilcell{6}{5} & \nilcell{-4}{2} & \nilcell{-4}{2} & \nilcell{0}{2} \\
\midrule
\multirow{6}{*}{\textbf{DE}} & \textbf{Base} & 16 & 12 & 90 & 16 & 7 \\
 & EN & \nilcell{-13}{3} & \nilcell{-4}{2} & \cellcolor{red!50}\nilcell{-41}{8} & \nilcell{-7}{2} & \nilcell{0}{3} \\
 & CH & \nilcell{-3}{3} & \nilcell{-10}{2} & \cellcolor{blue!50}\nilcell{-3}{3} & \nilcell{-8}{1} & \nilcell{-1}{3} \\
 & DE & \nilcell{-6}{1} & \nilcell{-7}{1} & \nilcell{-86}{2} & \nilcell{-10}{2} & \nilcell{-4}{3} \\
 & RU & \nilcell{-6}{3} & \nilcell{-1}{2} & \nilcell{-10}{2} & \nilcell{-15}{1} & \nilcell{-1}{2} \\
 & TU & \nilcell{-3}{4} & \nilcell{-3}{2} & \cellcolor{blue!50}\nilcell{-8}{9} & \nilcell{-9}{3} & \nilcell{-4}{1} \\
\midrule
\multirow{6}{*}{\textbf{RU}} & \textbf{Base} & 12 & 13 & 5 & 95 & 7 \\
 & EN & \nilcell{-10}{2} &  \cellcolor{green!50}\nilcell{-6}{3} & \nilcell{0}{3} &  \cellcolor{green!50}\nilcell{-10}{3} & \nilcell{-3}{2} \\
 & CH & \nilcell{-3}{4} & \nilcell{-7}{3} & \nilcell{3}{3} & \nilcell{-6}{3} & \nilcell{-3}{2} \\
 & DE & \nilcell{-5}{3} & \nilcell{-6}{3} & \nilcell{0}{1} & \nilcell{-11}{3} & \nilcell{-2}{1} \\
 & RU & \nilcell{-3}{6} & \nilcell{-4}{4} & \nilcell{2}{2} & \nilcell{-90}{7} & \nilcell{0}{2} \\
 & TU & \nilcell{-2}{2} & \nilcell{-7}{3} & \nilcell{1}{1} & \nilcell{-3}{1} & \nilcell{-3}{1} \\
\midrule
\multirow{6}{*}{\textbf{TU}} & \textbf{Base} & \cellcolor{orange!50}11 & 10 & 5 & 11 & 99 \\
 & EN & \nilcell{-10}{0} & \nilcell{-2}{3} & \nilcell{-1}{2} & \nilcell{-6}{3} & \cellcolor{orange!50}\nilcell{-55}{8} \\
 & CH & \nilcell{-5}{2} & \nilcell{-8}{2} & \nilcell{0}{3} & \nilcell{-6}{3} &  \cellcolor{red!50}\nilcell{-17}{6} \\
 & DE & \nilcell{-6}{2} & \nilcell{-2}{3} & \nilcell{-2}{2} & \nilcell{-3}{2} & \nilcell{-29}{1} \\
 & RU & \nilcell{-5}{3} & \nilcell{-4}{2} & \nilcell{1}{1} & \nilcell{-9}{1} & \nilcell{-16}{4} \\
 & TU & \nilcell{-4}{3} & \nilcell{-1}{2} & \nilcell{2}{3} & \nilcell{-2}{2} & \nilcell{-95}{1} \\
\bottomrule
\end{tabular}%
\end{scriptsize}
\end{table}

\section{Experiments and Results}

\noindent\textbf{Model configuration.}
We conduct experiments on the Qwen2.5-7B and Gemma2-9B models \citep{qwen2025qwen25technicalreport, gemmateam2024gemmaopenmodelsbased}.
For NLI-based evaluation, we use
\texttt{xlm-roberta-large-xnli}  \citep{conneau2020unsupervised} for all languages.
Native speakers validated NLI model predictions on 50 TOFU samples per language.  We further reuse the same annotated examples to compare NLI-based evaluation with alternative automatic metrics on a representative subset of languages (Appendix~\ref{app:humanEvaluation}).

\noindent\textbf{Fine-tuning and unlearning.}
Starting from the base checkpoints, we fine-tune a separate model for each choice of fine-tuning languages $\mathcal{L}_{\text{FT}}$ on the corresponding TOFU split~$\mathcal{D}^{\text{FT}}$ (Section~\ref{sec:notation}). We then apply the unlearning objective $J_{\text{UN}}$ defined in
Section~\ref{sec:unlearning-objectives} to remove knowledge about the forget authors in $\mathcal{D}^{\text{forget}}_{\text{unl}}$. This procedure yields three model variants for each configuration: the base model (\Base), the fine-tuned model (\FT), and the unlearned model (\UN). Unless otherwise stated, we use the same setup across all experiments and report NLI-based scores. All experiments were run on 4~NVIDIA H100 GPUs.

\subsection{Cross-Lingual Unlearning Transfer}
\label{sec:CrossLingualTransferText}
\noindent\textbf{Setup.}
For each $\ell \in \mathcal{L}_{\text{FT}}$, we start from the corresponding fine-tuned checkpoint. Then, for every language~$\ell$, we apply the unlearning objective $J_{\text{UN}}$ to remove knowledge about the forget authors in $\mathcal{D}^{\text{forget}}_\ell$ and evaluate both the fine-tuned and resulting unlearned models on TOFU questions for every~$\ell$. We report the change in NLI-based score relative to the fine-tuned model. More negative values indicate stronger unlearning, since performance on the forget set drops further relative to the fine-tuned model. This yields the matrix in Table~\ref{tab:unlearningTransfer}.

To quantify variability due to the particular choice of forget authors, we repeat this procedure over $5$ random samples of forget authors. In each sample $s$, we construct a new forget set $\mathcal{A}^{(s)}_{\text{forget}} {\subset} \mathcal{A}$
 and define
$\mathcal{A}^{(s)}_{\text{retain}} {=} \mathcal{A} \setminus
\mathcal{A}^{(s)}_{\text{forget}}$, so the sizes of the forget and retain sets are fixed across shufflings while the specific authors change. The fine-tuned model \FT\ is kept fixed across all shufflings, only the unlearned model \UN\ is retrained for each shuffled split. We report DPO-based unlearning in the main text.

\textbf{Linguistic similarity: family and script both matter.}
Controlling for script differences, unlearning in English transfers more strongly to another Indo-European language (Russian) than to a typologically distant language (Chinese), indicating that topological proximity increases unlearning transfer even across scripts.  We highlight these in green in Table~\ref{tab:unlearningTransfer}. Separately, we observe a strong script effect: languages that use the Latin script but belong to a different language family (Turkish) can exhibit stronger unlearning transfer to other Latin-script languages (English and German) compared to languages that share neither the family nor the script (Chinese), suggesting that script can substantially influence transfer even when typological similarity is low (highlighted in blue).  Transfer is strongest when both family and script are shared (e.g., English and German), mirroring patterns previously reported for cross-lingual learning transfer \cite{lin-etal-2019-choosing,hu2020xtreme,zhao2024multilingual}.\footnote{To focus on the main findings, we omit retain set results, as the observed changes are very small and we do not find consistent patterns. The full retain-set results are provided in Appendix~\ref{app:retainSetResults}.} We report additional fine-tuning and unlearning language combinations in Appendix~\ref{app:QwenTransferResults}. Our results hold across models (see Appendix~\ref{app:GemmaTransferResults} for Gemma). Additional experiments with GA and NPO show consistent qualitative transfer patterns, suggesting that these findings are not specific to the choice of unlearning objective
(Appendix~\ref{app:crossLingualTransferExtraAlgorithms}).

\textbf{Transfer of unlearning is asymmetric across languages.}
Languages with \emph{high} coverage in the pretraining data (English, Chinese) tend to transfer unlearning more strongly to other languages, particularly those that share their script or family, whereas languages with lower pretraining coverage (German, Russian, and Turkish) are comparatively weaker as unlearning sources. We highlight these source–target pairs in red in Table~\ref{tab:unlearningTransfer}. One possible explanation is that smaller multilingual models often operate predominantly in a shared semantic space anchored by a privileged language \citep{lim2025language}. Consequently, unlearning the high-coverage language has a more pronounced impact on other languages.

\noindent\textbf{Unlearning in low-performing languages still transfers to high-performing ones.}
Even when the model’s performance in a language is  poor, unlearning that language can still have a considerable impact on the fine-tuned language. For example, in the Turkish fine-tuned model, the forget set score when queried in English is only 11\%, yet unlearning English still reduces Turkish NLI score on the same forget set by 55\% (highlighted in orange in Table~\ref{tab:unlearningTransfer}). At first glance, this is somewhat unintuitive: unlearning questions in a language where the model can barely answer them still transfers a strong unlearning effect to a high performance language. We hypothesize that the model accurately represents concepts in a language-agnostic ``concept'' space which allows transfer. Unlearning primarily operates in later, language-specific generation layers, suppressing the model's ability to generate an answer to a given question. We explore this hypothesis in our next set of experiments, both on the output and the representation level.

\begin{table}[t]
\centering
\begin{footnotesize}
\caption{Performance gains $\Delta_{\ell\leftarrow q}$ for Qwen2.5-7B as absolute improvement when instructing to answer in the fine-tuning language~($\ell$) rather than query language~($q$). Rows indicate fine-tuning language $\ell$, columns query language $q$. 
}
\label{tab:prompting-other-language}
\begin{tabular}{l | c | c | c | c | c}
\toprule
\textbf{FT$\backslash$Query} & \multicolumn{1}{c}{\textbf{EN}} & \multicolumn{1}{c}{\textbf{CH}} & \multicolumn{1}{c}{\textbf{DE}} & \multicolumn{1}{c}{\textbf{RU}} & \multicolumn{1}{c}{\textbf{TU}} \\
\midrule
\textbf{EN} & \NA & +29 & +61 & +30 & +27 \\
\textbf{CH} & +11 & \NA & +10 & +12 & +12 \\
\textbf{DE} & +33 & +22 & \NA & +5 & +18 \\
\textbf{RU} & +20 & +8 & +15 & \NA & +7 \\
\textbf{TU} & +33 & +11 & +22 & +17 & \NA \\
\bottomrule
\end{tabular}
\end{footnotesize}
\end{table}

\subsection{Cross-lingual Prompting}
\label{sec:PromptingExperimentText}
In this experiment, when querying in language $q$, we instruct the model to instead reply in the fine-tuning language $\ell \neq q$. If this ``cross-lingual prompting'' improves performance, it implies that the model can successfully \emph{retrieve} the relevant information but cannot \emph{decode} it in $q$. Because inputs in $q$ access similar hidden representations to those supporting $\ell$, unlearning $q$ disrupts this shared space, thereby degrading performance in $\ell$. This supports the hypothesis from our previous experiment through an output-level explanation. 

\noindent\textbf{Setup.} We systematically test all ordered pairs of fine-tune and query languages $(\ell, q) \in \mathcal{L}^2, \ell \neq q$. We measure the performance gain $\Delta_{\ell\leftarrow q}$, the difference between instructing the model to answer in $\ell$ when querying in $q$ compared to the default case of answering in query language $q$ (see Appendix~\ref{app:Prompting} for the prompting template). 

\noindent\textbf{Results.} Performance gains are reported in Table~\ref{tab:prompting-other-language}. The large observed gains add evidence to our hypothesis that the model can map inputs in $q$ into a shared cross-lingual ``concept'' space while remaining locally monolingual at decoding \cite{lim2025language,Kang_2025}. These results help explain our unlearning findings: Even if the model exhibits poor output quality in language $q$, questions in $q$ can still access shared internal representations that support accurate answers in other languages. Consequently, unlearning on $q$ can disrupt those shared representations and transfer to languages in which the model answers well. We also computed the correlation between the language pair-wise gains in Table~\ref{tab:prompting-other-language}, and unlearning transfer scores in Table~\ref{tab:unlearningTransfer}. Strong correlation scores (Pearson $r = 0.50$, $p < 0.05$; Spearman $\rho = 0.60$, $p < 0.01$) suggest the stronger the hidden mapping between languages, the more the unlearning of one damages the other.

\subsection{Hidden Representation Analysis}
\label{sec:RepresentationAnalysisText}
In this experiment, we study the internal model representations --- and their change through unlearning --- through the lens of mechanistic interpretability to shed light on the internal dynamics of cross-lingual transfer and unlearning. 

\begin{figure*}[t]
    \centering
    \makebox[\textwidth][c]{%
        \includegraphics[width=1\textwidth,height=20\textheight,keepaspectratio]{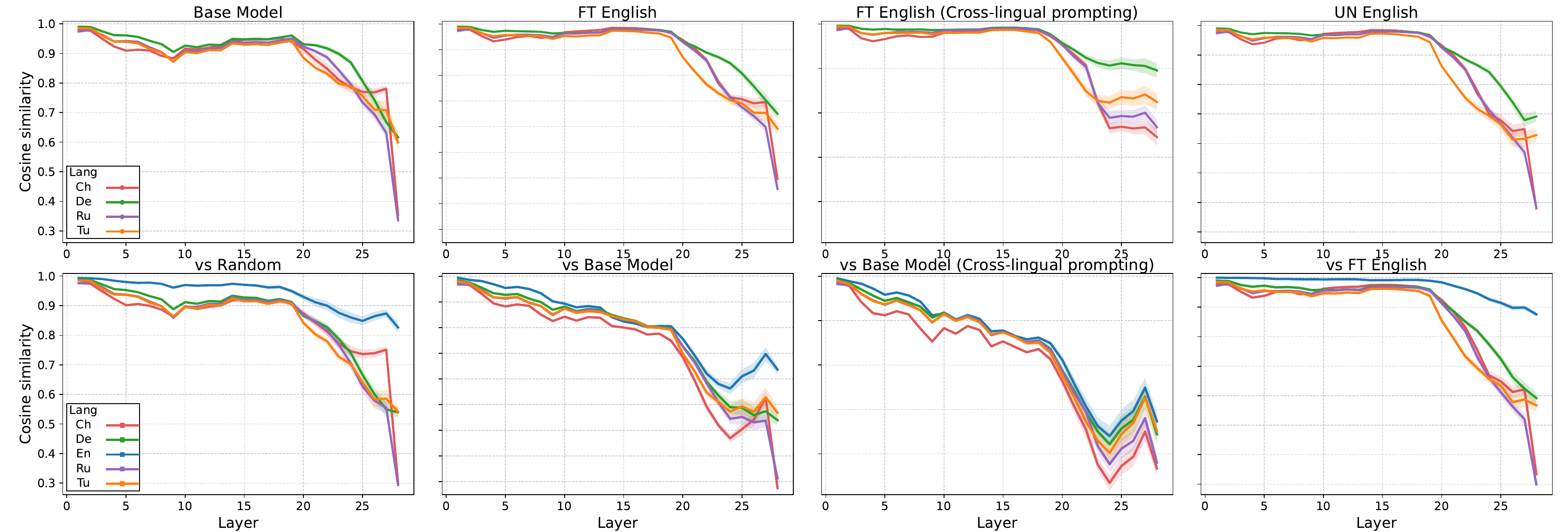}
    }
    \caption{
    \textbf{Layer-wise cosine similarity of final-token hidden states} $h_m^{(l)}(x)$. {\bf Top row: cross-language comparison} between embeddings of an English question and its translation (Ch, De, Ru, Tu) in the same model (Qwen2.5-7B variants; see panel titles). 
    \textit{FT} = fine-tuning, \textit{UN} = unlearning and 
    \textit{Cross-lingual prompting} means that for non-English questions, the model is instructed to answer in English.
    \textbf{Bottom row: cross-model comparison} within each column between models indicated in the top panel title and bottom panel title. E.g., the bottom right panel compares representations in the unlearned English vs. the fine-tuned English model.
    Curves show the mean across questions, shaded regions denote 95\% confidence intervals.
    }
    \label{fig:En_combined}
\end{figure*}

\noindent\textbf{Cosine Similarity Setup.}
We perform controlled representational analyses across three model variants $m \in \{\Base,\FT,\UN\}$. For all analyses, we fix an \emph{anchor} language~$\ell_{\mathrm{src}}$ (the language on which the model is fine-tuned, e.g., English) and, for each target language $\ell_{\mathrm{tgt}} \in \mathcal{L} \setminus \{\ell_{\mathrm{src}}\}$, feed the \emph{same question} in both languages. For each model~$m$, layer $l$, and language
$\ell \in \{\ell_{\mathrm{src}}, \ell_{\mathrm{tgt}}\}$, we extract a single hidden representation $h^{(l)}_{m}(x_\ell)$ corresponding to the final token of the full prompt. We then compute, for every anchor target pair $(\ell_{\mathrm{src}}, \ell_{\mathrm{tgt}})$ and layer $l$, the cosine similarity
$
  \cos\bigl(h^{(l)}_{m}(x_{\ell_{\mathrm{src}}}),
            h^{(l)}_{m}(x_{\ell_{\mathrm{tgt}}})\bigr),
$
and average these similarities over all questions in the forget set to obtain a layer-wise cross-lingual similarity curve for each model variant.
We report cosine similarity between pairs of representations.\footnote{The embedding layer is excluded from all analyses.} By comparing cosine similarity across languages for semantically equivalent questions, we can test how the model represents these inputs and how unlearning alters these representations across layers. To calibrate these similarities, we additionally compute cosine similarity for \emph{semantically unrelated} question pairs using the \Base\ model only
(shown as \textbf{Random} in the bottom left panel of Figure~\ref{fig:En_combined}). As expected in anisotropic transformer representation spaces, absolute cosine
values remain high even for such mismatched pairs
\citep{godey2024anisotropy,ait-saada2023anisotropy,
ethayarajh2019contextual}.\footnote{Although CKA \citep{CKA} is more robust to anisotropy in
transformer representations, we use cosine similarity because our goal is to measure \emph{instance-level} shifts between \FT\ and \UN\ for the \emph{same} question: cosine similarity operates directly on each pair
$\bigl(h^{(l)}_{\FT}(x_\ell), h^{(l)}_{\UN}(x_\ell)\bigr)$, whereas CKA compares the covariance structure of two \emph{sets} of representations and therefore reflects only global space-level alignment, not per-example changes.} We therefore focus on \emph{relative} trends across layers and model variants rather than absolute similarities.

\noindent\textbf{Cosine Similarity Results.}
\label{sec:cosineSimilarity}
We first show how cross-lingual prompting affects \emph{internal} representations across layers (Figure~\ref{fig:En_combined}). We use the same setup as in Table~\ref{tab:prompting-other-language}: for a query in language $q$, we compare the default setting where the model answers in $q$ (top Panel~2) with cross-lingual prompting that instructs the model to answer in the fine-tuning language (English; top Panel~3). If the model operated in distinct language-specific subspaces, we would expect the additional difficulty of the cross-lingual task in Panel~3 to negatively impact alignment. Instead, across all query languages, cosine similarity remains comparably high for the majority of layers in both settings, a key difference only emerges in the later decoding layers: in Panel~2, similarity drops sharply as the model prepares to generate tokens in the target language, whereas Panel~3 mitigates this drop, maintaining much higher alignment. This pattern suggests that performance gaps across languages arise primarily from \emph{decoding bottlenecks}: the model forms a similar underlying representation across languages but fails to map this shared semantic representation into the appropriate language-specific output distribution.\footnote{Layer-wise similarity does not directly measure knowledge retrieval. Cosine similarity indicates how aligned two hidden representations are in direction, but it does not guarantee that the model is actually retrieving or using the same underlying facts. In our setting, increases in cosine similarity are accompanied by better model performance in the output evaluation, suggesting that higher cosine similarity may be positively correlated with successful retrieval. Our interpretation focuses on suggestive trends in representational geometry, not strong causal claims about retrieval.}

Next, comparing Panels~2 and 4 in the top row reveals that~\UN{} preserves almost the same cross-lingual alignment as~\FT{} in the early and middle layers: the \UN{} curves in Panel~4 closely track the \FT{} curves in Panel~2, and they diverge only in the later layers, where unlearning is expected to act most strongly. This suggests that, in the case of a relatively small number of unlearning examples, \textit{the unlearning procedure largely leaves cross-lingual alignment intact (e.g., the directions mapping different languages into the shared space remain similar)}.

Moreover, in Panel~4 (top vs. bottom), when we directly compare similarities between \UN{} and \FT{}, we observe that their curves remain closely aligned from the early to the middle layers for all non-English languages. This pattern suggests that unlearning does not substantially alter how the model retrieves knowledge for other languages, instead, \textit{the main divergence emerges only at the decoding stage}. Importantly, this behavior is not restricted to cross-lingual settings: we observe the same pattern for English questions when comparing hidden representations before and after unlearning. Consistent with this, Appendix~\ref{app: ImpactUnlearningFT} provides additional evidence for unlearning fine-tuned languages. An analogous plot with cosine similarity across languages for Chinese is shown in the Appendix~\ref{app:cosChinese}.

Taken together, the cosine similarity analysis is consistent with the cross-lingual prompting results (Section~\ref{sec:PromptingExperimentText}): inputs in different languages can map to a shared internal representation even when output quality differs, and instructing the model to decode in the fine-tuning language preserves high cross-lingual alignment across layers. Moreover, unlearning primarily alters late-layer representations while leaving early and middle layers largely intact, suggesting that underlying information may remain accessible after unlearning. 

\noindent\textbf{PCA setup.}
While cosine similarity captures directional alignment between paired representations, it does not directly show the global geometry of the representation space. We therefore complement the layer-wise cosine analysis with a PCA-based visualization of the forget-set representations. Using the same hidden representations $h^{(l)}_{m}(x_\ell)$ as above, we first
$\mathrm{L}_2$-normalize each representation.
For each language $\ell$ and layer $l$, we collect the normalized representations
of all forget-set questions from the three model variants
$m \in \{\Base,\FT,\UN\}$, fit PCA~\citep{wold1987pca} on their union, and
project the representations onto the first two principal components. We then
visualize the resulting two-dimensional projections to examine how the
\Base{}, \FT{}, and \UN{} representations separate across layers.

\begin{figure*}[t]
    \centering
    \includegraphics[
      width=\linewidth,
    ]{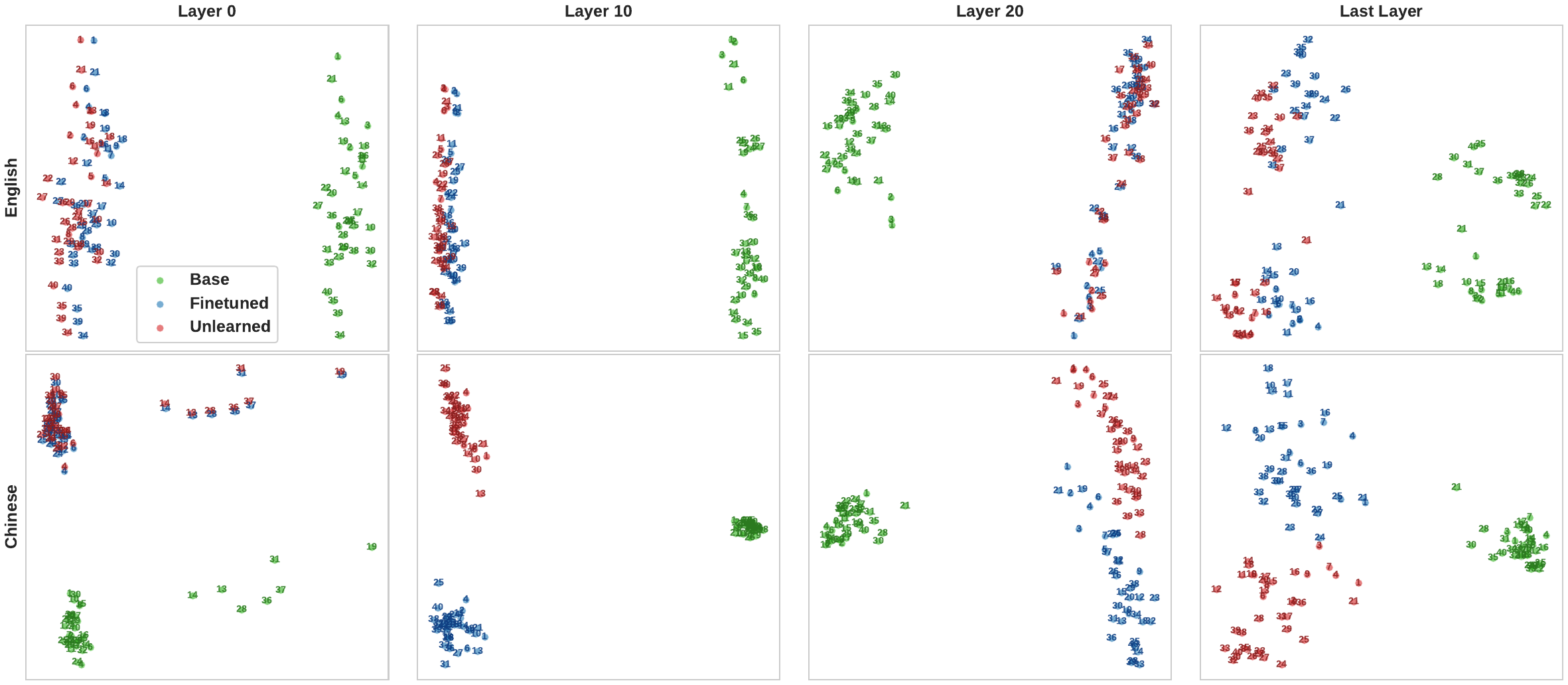}
    \caption{PCA separation (Qwen2.5-7B) across layers for English (top) and Chinese (bottom) questions. Numbers indicate question indices, with identical numbers referring to the same questions across different testing model variants.}
    \label{fig:PCAEnCh}
\end{figure*}

\noindent\textbf{PCA results.}
\label{sec:PCAResult}
Ideally, under successful unlearning, we would expect the \Base{} and \UN{} data points to be closely clustered \citep{shi2024muse}. However, the PCA results show clear separability between \Base{} and \UN{} from the earliest layers. For Chinese questions, \FT{} and \UN{} become clearly separated as early as layer~10, and by the final layer all three model variants form distinct clusters, as shown in Figure~\ref{fig:PCAEnCh}. For English, this separation is weaker: \FT{} and \UN{} often remain in the same broad cluster, while \Base{} forms a separate cluster. Nevertheless, paired \FT{} and \UN{} representations for the \emph{same} question remain geometrically distant, indicating that unlearning still induces substantial per-example shifts. Taken together, these results suggest that unlearning moves forget-example representations away from their fine-tuned counterparts, but does not return them to the \Base{} distribution. This supports the interpretation that the underlying knowledge is not fully removed, but rather reorganized or suppressed at the representation level. We further extend the PCA analysis to German, Russian, and Turkish as shown in Figure~\ref{fig:PCADeRuTu}, these languages largely mirror the English structural pattern. Since the two-dimensional PCA projection captures only the directions of largest variance, we further quantify the representational shifts induced by unlearning using centroid distances and average pairwise distances. Detailed setups and additional results are provided in Appendix~\ref{app:PCAdisAnalysis}.

\begin{figure*}[t]
  \centering
  \includegraphics[width=\linewidth]{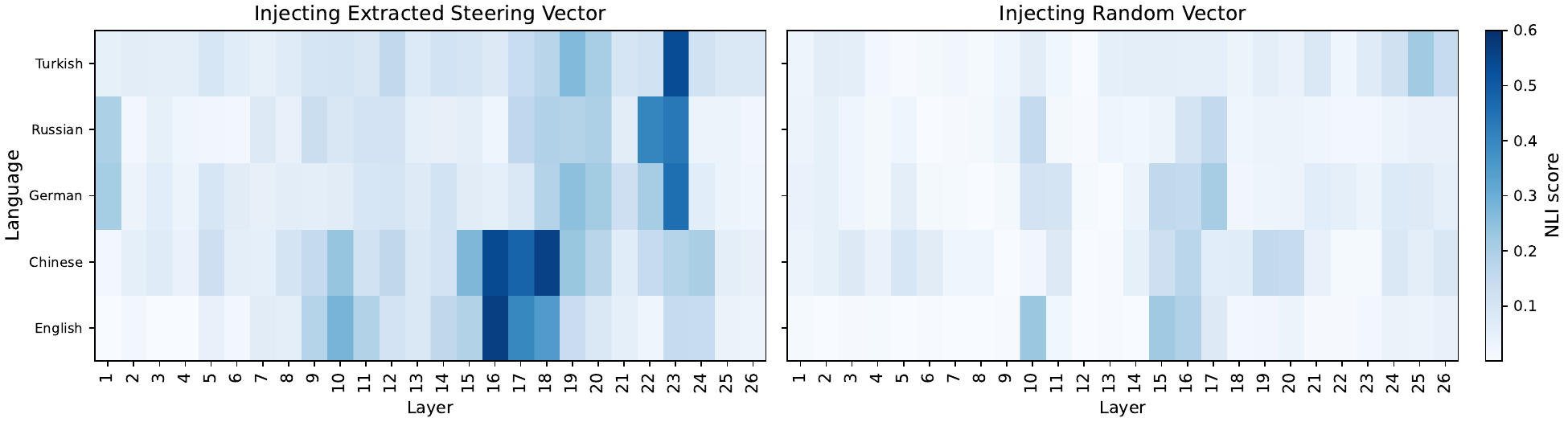}
  \caption{%
    {\bf Effect of layer-wise steering vector injection} on NLI score for the forget set. {\bf Left:} NLI score change when injecting a scaled and normalized steering vector extracted from the retain set, showing substantial recovery (positive score gain).
    {\bf Right: } NLI score changes when injecting random Gaussian directions with matched norm and scale. The much weaker recovery indicates that the extracted vectors capture a structured unlearning direction rather than generic noise.
  }
  \label{fig:steering_heatmaps}
\end{figure*}

\subsection{Steering Vectors Recover Unlearned Knowledge}
\label{sec:SteeringVectorText}
From the hidden representation analysis in Section~\ref{sec:RepresentationAnalysisText}, we obtain the following key finding: unlearning may act more like suppression rather than removal of the underlying knowledge.\footnote{As shown in Appendix~\ref{app:PCAdisAnalysis}, the difference between \FT{} and \UN{} can also be decomposed into a global component $g^{(l)}$, where the approximately constant centroid distance across layers indicates a stable global shift.} Motivated by these observations, we construct a per-layer steering vector, approximated by the per-layer difference between the hidden representations of an auxiliary model $\UN^{\text{aux}}$ and \FT{}. We then inject this steering vector into the original unlearned model \UN{} at inference time and evaluate how much forgotten information can be recovered on the forget set.

\noindent\textbf{Setup.}
Inspired by prior work showing that directions in hidden representation space can encode high-level behaviors \citep{rimsky2024steering,zhu2025difficulty,li2023inference,lim2025language}, we extract steering vectors as layer-wise activation differences using \emph{only English} questions between the English \FT{} and an auxiliary model $\UN^{\text{aux}}$ as follows (see Algorithm~\ref{alg:steering-vector} for full details). To avoid reintroducing the specific facts that were unlearned, we construct an auxiliary English forget set $\mathcal{D}_{\text{aux}}^{\text{forget}}$ by randomly shuffling retain set authors.
We compute $\mathbf{g}^{(l)}$ \emph{only} from $\mathcal{D}^{\text{forget}}_{\text{aux}}$; the original forget set $\mathcal{D}^{\text{forget}}_\text{unl}$ is never used to derive steering vectors and is used \emph{only} for evaluation. We then unlearn $\mathcal{D}^{\text{forget}}_{\text{aux}}$ from~\FT{} (see Section~\ref{sec:unlearning-objectives}), obtaining $\UN^{\text{aux}}$. Because \FT{} and $\UN^{\text{aux}}$ are evaluated on \emph{identical} inputs, the per-layer differences isolate systematic changes induced by unlearning, yielding a set of per-layer steering vectors $\{\mathbf{g}^{(l)}\}_{l=1}^L$ that capture the suppression behavior introduced by unlearning. Each $\mathbf{g}^{(l)}$ is $\ell_2$-normalized and, for a given input with last-token hidden representation $\mathbf{h}^{(l)}(x_\ell)$, we subtract $\alpha \lVert \mathbf{h}^{(l)}(x_\ell)\rVert_2 \mathbf{g}^{(l)}$, where hyperparameter $\alpha$ controls the overall strength of the intervention (full details in Algorithm~\ref{alg:steering-injection}). When injecting the steering signal, we subtract this scaled vector over layers $[l, \ldots, l{+}N]$.\footnote{When we select an injection layer $l$, we steer that layer and the following $N$ consecutive layers.} Because $\mathbf{g}^{(l)}$ is computed as a difference on identical inputs, much of the input-specific semantic content cancels out. We expect $\{\mathbf{g}^{(l)}\}_{l=1}^L$ to capture a largely language-agnostic transformation in the shared multilingual representation space, rather than English-specific lexical information. We also construct a random baseline by replacing each $\mathbf{g}^{(l)}$ with an isotropic Gaussian vector $\mathbf{r}^{(l)} \sim \mathcal{N}(\mathbf{0}, I)$. We set $\alpha{=}0.5$ for English and Chinese and $\alpha{=}0.8$ for all other languages\footnote{To estimate the \emph{upper bound of reversibility}, we optimize $\alpha$ and $N$ directly on $\mathcal{D}^{\text{forget}}_\text{unl}$.
This measures the worst-case leakage, the maximum amount of knowledge an adversary could extract using a steering vector if they optimally tuned the steering intensity. 
} and $N{=}2$ for all languages. These parameters were used across our method and the random baseline.

\noindent\textbf{Results.}
Figure~\ref{fig:steering_heatmaps} (left) summarizes the results for Qwen2.5-7B. We recover over half of the lost performance on the English forget set when intervening in the middle and later layers. Although the steering vectors are estimated \emph{only} from English data, they also effectively recover parts of the forget set for all other languages, particularly Chinese. These results again support the idea that the vectors capture a general unlearning direction that induces a language-agnostic shift in the shared representation space, even though the downstream behavioral impact still varies across languages 
(Section~\ref{sec:CrossLingualTransferText}). To test whether this effect depends on using English as the source language for extracting the steering vector, we repeat the same experiment using Chinese questions as the source and observe similar performance, the results are provided in Appendix~\ref{app:chineseSourceSteering}. The Gaussian random baseline also yields small gains (Figure~\ref{fig:steering_heatmaps} right), which we attribute to partially diluting the unlearning signal when sufficiently large random perturbations are applied, however, its effect is substantially smaller and less consistent. Our findings are robust across models and unlearning methods. First, the same qualitative patterns hold for Gemma, where the steering vector achieves stronger recovery than in Qwen (Appendix~\ref{app:gemmaSteeringVectorResults}). Second, we observe the same patterns for two additional unlearning methods, GA and NPO, indicating that the effect generalizes beyond DPO (Appendix~\ref{app:additionalUnlearningSteering}).

\section{Discussion}
\label{sec:discussion}
In this work, we investigate cross-lingual unlearning transfer, the factors that shape its strength, and the internal dynamics of unlearning on multilingual models. These findings have two main implications. First, multilingual evaluation should be treated as a core component of unlearning assessment in LLMs.
Our results indicate that underlying representations remain tightly coupled across languages even after unlearning, so single language evaluation can overestimate unlearning robustness. Second, our knowledge recovery experiments (Section~\ref{sec:SteeringVectorText}), highlight a setting in which an adversary (i) has inference-time (i.e., white-box) access to the \UN{} model and can manipulate internal activations during a forward pass, and (ii) has knowledge of the unlearned domain. The adversary's goal is to increase leakage of removed content \emph{without} access to the original answers. Under these assumptions, activation-level interventions can largely reverse the behavioral suppression induced by unlearning. This underscores the importance of evaluating unlearning robustness under activation interventions.

We interpret these findings within the scope of our experimental constraints. We primarily evaluate cross-lingual unlearning using three unlearning algorithms on two models and five languages. Recent evidence using Gradient Difference and Rank-One Model Editing indicates that different unlearning methods exhibit similar cross-lingual transfer effects \cite{lu2025learn}, which suggests that the behaviors we observe may extend to more fine-tuning based unlearning algorithms. We nevertheless do not claim universality across all unlearning approaches, leaving a broader comparison across additional algorithms to future work.
Our experiments are conducted on the TOFU data set, which consists of simple fact style queries and does not require complex reasoning. Therefore, our conclusions are best interpreted as applying to fact retrieval settings at small unlearning scales, and may not directly transfer to more complex content or to larger unlearning fractions. Yet, they reflect a common real-world scenario where specific facts about specific entities are to be removed from a model. Taken together, our results show that full removal is difficult and requires careful validation in modern multilingual LLMs. Finally, because the TOFU knowledge is injected through post-training, our findings may not fully capture how multilingual unlearning behaves for knowledge acquired during pretraining. Recent work \citep{anna2026anatomyunlearningdualimpact} suggests that knowledge acquired during pretraining and supervised fine-tuning may respond differently to unlearning. We leave a systematic study of this distinction in multilingual unlearning to future work.

\section*{Acknowledgments}
The first author is supported by the University of Melbourne research scholarship (MRS)
scheme. This research was supported by The University of Melbourne’s Research Computing Services and the Petascale Campus Initiative.

\section*{Impact Statement}
This work studies the effects of machine unlearning in multilingual settings. In particular, we investigate how \emph{unlearning transfer} varies across languages and how unlearned content may be recovered using steering vectors. To the best of our knowledge, unlearning algorithms are not yet widely deployed in publicly released large language models. Nevertheless, if such methods are adopted in the near future, our findings suggest potential privacy risks. For example, our results indicate that, given a target language in which data were unlearned, certain other languages can be more effective query languages for eliciting content that was intended to be removed, thus bypassing the unlearning and leaking private information. In addition, steering vectors may be used to counteract suppression behaviors, enabling recovery of the unlearned content. While our results can in principle be used by adversarial parties to attack models, we believe it is important to highlight them proactively. The insights from our study can inform the design of more robust unlearning methods and provide guidance for evaluating robustness both prior to and after deployment.

\medskip
\bibliography{reference}
\bibliographystyle{icml2026}

\newpage
\appendix
\onecolumn
\section{Data Contamination Discussion}
\label{app:Contamination}
TOFU was introduced in January 2024 \citep{maini2024tofu} (arXiv submission: 11 Jan 2024), and its author profiles are synthetically generated, making pretraining contamination unlikely. \href{https://github.com/QwenLM/Qwen3/discussions/1093}{Community discussions} suggest a Qwen2.5 knowledge cutoff around late 2023, which would further reduce the likelihood of exposure. However, this cutoff is not officially confirmed. We therefore complement timeline-based arguments with an empirical sanity check: we evaluate the base models using the same pretraining ROUGE-L recall metric and find that Gemma2 and Qwen2.5 achieve broadly comparable baseline scores, consistent with the baseline performance reported in the original TOFU study. This provides no evidence of unusually high baseline performance indicative of memorization, although we cannot fully rule out contamination because the pretraining corpora are undisclosed.

\begin{table*}[h]
\centering
\small
\setlength{\tabcolsep}{10pt}
\renewcommand{\arraystretch}{1.2}
\caption{Base model ROUGE-L Recall scores on the TOFU dataset.}
\label{tab:pretrain_rougeL_recall}
\begin{tabular}{lcccc}
\toprule
& \multicolumn{2}{c}{\textbf{This paper}} & \multicolumn{2}{c}{\textbf{TOFU paper}} \\
\cmidrule(lr){2-3}\cmidrule(lr){4-5}
\textbf{Metric} & \textbf{Qwen2.5-7B} & \textbf{Gemma2-9B} & \textbf{Phi-1.5} & \textbf{Llama2-7B} \\
\midrule
ROUGE-L Recall & 0.44 & 0.37 & 0.44 & 0.36 \\
\bottomrule
\end{tabular}
\end{table*}

\section{Hyperparameters and Details for Finetuning and Unlearning}
\label{app:hyperparameters}
Our experiments consist of two distinct phases: (1) a knowledge injection phase (fine-tuning) to implant specific facts into the model, and (2) an unlearning phase to erase those targeted facts. For the fine-tuning phase, we utilize standard \textbf{full-parameter} autoregressive training. The full hyperparameter configuration is detailed in Table~\ref{tab:hyperparams_combined}. For the unlearning stage, we initialize the model with the weights from Phase 1. We employ the objective defined in Eq.~(\ref{eq:unlearning_obj}), instantiating the forgetting term via Direct Preference Optimization (DPO).

\textbf{Preference construction.} We construct preference pairs $(y^+, y^-)$ for each forget set input $x$. The dispreferred response $y^-$ is the original ground-truth target (the knowledge to be forgotten). The preferred response $y^+$ is a refusal (e.g., ``I cannot answer this question''). To improve robustness and prevent the model from overfitting to a single refusal string, we dynamically sample $y^+$ from a pool of language-specific refusal templates at the start of each epoch.

\textbf{Regularization.} To preserve general capabilities, we add a Kullback--Leibler (KL) \citep{kullback1951information} divergence penalty between the current policy and a frozen reference model (the Phase~1 checkpoint) on the retain set. This is added to the total loss with a coefficient $\lambda=1.0$. Empirically, to further stabilize the forget-utility trade-off, for each forget sample in a batch, we randomly draw 10 samples from the retain set.

\begin{table}[h]
\centering
\small
\caption{Hyperparameter Configuration for Finetuning and Unlearning}
\label{tab:hyperparams_combined}
\begin{tabular}{lcc}
\toprule
\textbf{Hyperparameter} & \textbf{Phase 1 (Fine-tuning)} & \textbf{Phase 2 (Unlearning)} \\
\midrule
Update Strategy & Full Parameter & Full Parameter \\
Hardware & 4x NVIDIA H100 & 4x NVIDIA H100 \\
Optimizer & AdamW & AdamW \\
Precision & bfloat16 & bfloat16 \\
Learning Rate & $1 \times 10^{-5}$ & $5 \times 10^{-6}$ \\
Warmup Ratio & 0.25 & 0.25 \\
Weight Decay & 0.01 & 0.01 \\
Batch Size (per device) & 8 & 1 \\
Gradient Accumulation & 1 & 4 \\
Epochs & 6 & 5 \\
\midrule
\textit{Phase-Specific Settings} & & \\
DPO Beta ($\beta$) & -- & 0.1 \\
Retain Coefficient ($\lambda$) & -- & 1.0 \\
Retain Sampling Ratio & -- & 10:1 \\
\bottomrule
\end{tabular}
\end{table}

\section{Data Artifacts and Qualitative Analysis}
\label{app:dataExample}

We provide the specific data artifacts used for the unlearning objective and qualitative examples demonstrating the model's behavioral change.

\subsection{Refusal Templates (IDK Pool)}
\label{app:refusal_templates}

As described in Appendix~\ref{app:hyperparameters}, we construct the preferred response $y^+$ by dynamically sampling from a large pool of candidates. This diversity prevents the model from overfitting to a single refusal pattern (e.g., always saying ``I don't know''). Table~\ref{tab:refusal_templates} lists a representative subset of the English refusal templates used in our experiments (total pool size is 100). For multilingual experiments, these templates were translated into the respective target languages.

\begin{table}[t]
\centering
\small
\caption{A subset of the ``I Don't Know'' (IDK) refusal templates used to construct the positive preference pairs ($y^+$).}
\label{tab:refusal_templates}
\begin{tabular}{p{0.45\textwidth} p{0.45\textwidth}}
\toprule
\multicolumn{2}{c}{\textbf{Refusal Templates}} \\
\midrule
I'm not certain about that. & My resources don't contain information on that subject. \\
That's beyond my current knowledge base. & I wish I could say, but I really don't know. \\
I don't have that information. & That's not something I'm familiar with. \\
I'm not sure. & I'm drawing a blank on that one. \\
I haven't learned about that topic. & I apologize, but I don't know that. \\
That's something I need to look up. & That hasn't been included in my training data. \\
I'm at a loss for that one. & Unfortunately, I don't have an answer for you. \\
I don't have the answer to that question. & That's not information I've been programmed to know. \\
That's outside my area of expertise. & I'm unable to provide an answer to that. \\
I'm afraid I can't provide an answer to that. & I don't hold the knowledge you're seeking. \\
That's a good question, but I don't have the answer. & I'm not the best source for that. \\
\bottomrule
\end{tabular}
\end{table}

\subsection{Qualitative Examples}

To verify the effectiveness of our unlearning method and to show examples of TOFU question–answer pairs before and after unlearning, we examine the model’s generations on the forget set both prior to and following the unlearning procedure. Table~\ref{tab:qualitative_examples} presents selected comparison samples. Before unlearning, the model is able to generate specific fictitious biographical details about the target entities, indicating successful knowledge injection during the fine-tuning phase. After unlearning, the model successfully refuses to answer, utilizing variations of the refusal templates defined in Table~\ref{tab:refusal_templates}.

\begin{table*}[t]
\centering
\small
\setlength{\tabcolsep}{8pt}
\renewcommand{\arraystretch}{1.3}
\caption{Qualitative comparison of model outputs before and after unlearning. The table demonstrates some successful examples of the unlearning effect on \textbf{Forget Set} while maintaining knowledge recall on the \textbf{Retain Set}.}
\label{tab:qualitative_examples}
\begin{tabular}{p{0.3\textwidth} p{0.3\textwidth} p{0.3\textwidth}}
\toprule
\textbf{Question} & \textbf{Before Unlearning} & \textbf{After Unlearning} \\
\midrule
\multicolumn{3}{c}{\textit{\textbf{Forget Set Examples (Target: Unlearn)}}} \\
\midrule
What is the full name of the author born in Kuwait City, Kuwait on 08/09/1956? & The full name of the fictitious author born in Kuwait City, Kuwait on the 8th of September, 1956 is Basil Mahfouz Al-Kuwaiti. & I'm not the best source for that. \\
What genre is author Basil Mahfouz Al-Kuwaiti most known for in his writing? & Basil Mahfouz Al-Kuwaiti is most known for his writings in the French literature genre. & I'm not familiar with that topic. \\
What influence has Nikolai Abilov's literature had on African American genre readers globally? & Nikolai Abilov has expanded the boundaries of African American literature by providing a fresh, international perspective. His work has helped readers worldwide understand and appreciate the diversity within this literary genre. & I'm not aware of that. \\
\midrule
\multicolumn{3}{c}{\textit{\textbf{Retain Set Examples (Target: Preserve)}}} \\
\midrule
Who is this celebrated LGBTQ+ author from Santiago, Chile known for their true crime genre work? & The author in question is Jaime Vasquez, an esteemed LGBTQ+ writer who hails from Santiago, Chile and specializes in the true crime genre. & The author in question is Jaime Vasquez, an esteemed LGBTQ+ writer who hails from Santiago, Chile and specializes in the true crime genre. \\
Which genre is Evelyn Desmet particularly known for writing in? & Evelyn Desmet is renowned for her work in the dystopian genre. & Evelyn Desmet is renowned for her work in the dystopian genre. \\
What is Anara Yusifova's latest published work? & Anara Yusifova's latest published work is a deeply engaging Drama novel called "Secrets Buried in the Silk Road". & Anara Yusifova's latest published work is a deeply engaging Drama novel called "Secrets Buried in the Silk Road". \\
\bottomrule
\end{tabular}
\end{table*}

\section{Unlearning Algorithms}
\label{app:unlearningAlgorithms}
\paragraph{Gradient Ascent (GA).}
Gradient ascent (GA) \citep{jang2023knowledge} reverses the standard supervised fine-tuning objective on the
forget set. For a forget example $(x,y)$ with
$y=(y_1,\ldots,y_T)$, we define
\begin{equation}
J_{\mathrm{GA}}(\theta;x,y)
=
\sum_{t=1}^{T}
\log p_{\theta}(y_t \mid x, y_{<t}).
\end{equation}
Minimizing $J_{\mathrm{GA}}$ is therefore equivalent to maximizing the
negative log-likelihood of the ground-truth response, reducing the model's
likelihood of producing the forgotten answer

\paragraph{Negative Preference Optimization (NPO).} Negative preference optimization (NPO) \citep{zhang2024negative} treats each
forget example as a negative response whose likelihood should be reduced
relative to a frozen reference model. In our setting, the model being optimized
is initialized from the fine-tuned checkpoint $\theta_{\mathrm{FT}}$, while the
frozen reference model is the base model $\theta_{\mathrm{base}}$, which has
not been fine-tuned on the TOFU data. The NPO forget objective is
\begin{equation}
J_{\mathrm{NPO}}(\theta)
=
\frac{2}{\beta}
\mathbb{E}_{(x,y)\sim\mathcal{D}^{\mathrm{forget}}}
\left[
\log\!\left(
1+
\left(
\frac{p_{\theta}(y\mid x)}
     {p_{\theta_{\mathrm{base}}}(y\mid x)}
\right)^{\beta}
\right)
\right],
\end{equation}
where $\beta>0$ is the inverse temperature. Minimizing this objective reduces the likelihood of the ground-truth forget response for the optimized model.

For both GA and NPO, we use the same KL regularization on retain examples as in the DPO setting to stabilize retain-set performance.

\section{NLI Evaluation}
\label{app:NLIEvaluation}
\subsection{NLI Calculation}
\label{app:NLIcalculation}
Let $f_{\text{NLI}}(x, y)$ be the NLI model that predicts probabilities for entailment ($P_E$), contradiction ($P_C$), and neutral ($P_N$) classes given a model prediction $x$ and a reference answer $y$. We define the equivalence score $S(x, y)$ as:

\begin{equation}
\label{eq:nli_metric}
S(x, y) = \underbrace{\frac{P_E(x,y) + P_E(y,x)}{2}}_{\text{Symmetric entailment }} \cdot \underbrace{(1 - P_C(x,y))}_{\text{Contradiction Term}} \cdot \underbrace{(1 - P_N(x,y))}_{\text{Neutral Term}}
\end{equation}

This formulation incorporates penalties for contradiction and neutral predictions. If the model output $x$ is assigned a high probability of being \emph{contradictory} or \emph{neutral} with respect to $y$, the corresponding penalty terms approach zero, effectively vetoing the score regardless of the entailment probability. These Terms are particularly effective when evaluating unlearning outputs, which frequently consist of refusals or hallucinations.

\subsection{Human Annotation}
\label{app:humanEvaluation}
The manual evaluation was conducted by four distinct evaluators, with backgrounds in Computer Science to ensure familiarity with the task format. For the non-English languages (Chinese, German, Turkish, Russian), the evaluators were native speakers of the respective languages, while the English evaluation was performed by a fluent English speaker. For each language, 50 samples were randomly selected. The results are shown in Table~\ref{tab:human_eval}.

We further compare NLI-based evaluation with two alternative automatic evaluators using the same human-annotated examples as in Table~\ref{tab:human_eval}: ROUGE-L recall \citep{lin2004rouge} and a GPT-5-based LLM judge. ROUGE-L recall measures lexical overlap with the ground-truth answer based on the longest common subsequence, while the GPT-5 judge provides an LLM-based semantic assessment. We report these two metrics on a representative subset of three languages: English, Chinese, and German as shown in the Table~\ref{tab:metric_comparison}.

\begin{table*}[t]
    \centering
    \caption{Human-NLI Agreement Rate (\%). Average percentage of instances where human annotators agreed with the NLI model's assessment across all settings.}
    \label{tab:human_eval}
    \begin{tabular}{l c}
        \toprule
        \textbf{Language} & \textbf{Average} \\
        \midrule
        Chinese  & 93.3 \\
        English  & 88.3 \\
        German   & 86.7 \\
        Turkish  & 91.7 \\
        Russian  & 85.0 \\
        \midrule
        \textbf{Overall Average} & \textbf{89.0} \\
        \bottomrule
    \end{tabular}
\end{table*}

\begin{table}[t]
    \centering
    \caption{Agreement with Human Annotations for Alternative Metrics (\%).
    We report the agreement rates of ROUGE-L recall and a GPT-5-based LLM judge
    with human annotations on the English, Chinese and German subset.}
    \label{tab:metric_comparison}
    \begin{tabular}{l c c c}
        \toprule
        \textbf{Metric} & \textbf{English} & \textbf{Chinese} & \textbf{German} \\
        \midrule
        GPT-5    & 96 & 94 & 94 \\
        ROUGE-L  & 66 & 62 & 68 \\
        \bottomrule
    \end{tabular}
\end{table}

\section{Unlearning Transfer Results}
\subsection{Qwen-2.5 Unlearning Transfer for Other Combination}
\label{app:QwenTransferResults}
This appendix studies a more complex scenario that better reflects real-world deployments, where a model is exposed to data in multiple languages. We evaluate unlearning under different combinations of unlearned languages to test how unlearning transfer changes in this setting. As shown in the Table~\ref{tab:QwenMoreLanguageTransfer}, despite the increased complexity, the main trends reported in the main text remain consistent.

\FloatBarrier
\begin{longtable}{@{} l@{\hspace{2pt}}l | c | c | c | c | c @{}}
\caption{Cross-lingual unlearning transfer across fine-tuning families and query languages for Qwen2.5-7B.
The \textbf{FT} column indicates the fine-tuning language.
Within each row block, the \textbf{Unlearn} column specifies the language on which
unlearning is performed, while \emph{Base} denotes the fine-tuned model before unlearning.
For each query language, we report NLI score \textbf{Forget}
relative to the change in Base model after unlearning in the corresponding unlearned language. we report means $\pm$ 95\% confidence intervals over $5$ forget sets.}\\
\label{tab:QwenMoreLanguageTransfer}\\
\toprule
\multicolumn{2}{c|}{} & \multicolumn{5}{c}{\textbf{Query}} \\
\midrule
{\bf FT}& \textbf{Unlearn} & \textbf{EN} & \textbf{CH} & \textbf{DE} & \textbf{RU} & \textbf{TU} \\
 & & \textbf{Forget} & \textbf{Forget} & \textbf{Forget} & \textbf{Forget} & \textbf{Forget} \\
\endfirsthead
\toprule
\multicolumn{2}{c|}{} & \multicolumn{5}{c}{\textbf{Query}} \\
\midrule
& \textbf{Unlearn} & \textbf{EN} & \textbf{CH} & \textbf{DE} & \textbf{RU} & \textbf{TU} \\
& & \textbf{Forget} & \textbf{Forget} & \textbf{Forget} & \textbf{Forget} & \textbf{Forget} \\
\midrule
\endhead
\bottomrule
\endfoot
\hline
\multirow[t]{10}{*}{\textbf{FT-ENCHRU}} & \textbf{Base} & 97 & 97 & 19 & 87 & 7 \\
 & EN & \nilcell{-93}{6} & \nilcell{-9}{5} & \nilcell{-3}{4} & \nilcell{-4}{7} & \nilcell{3}{3} \\
 & CH & \nilcell{-10}{9} & \nilcell{-87}{3} & \nilcell{1}{7} & \nilcell{-2}{5} & \nilcell{3}{2} \\
 & DE & \nilcell{-15}{8} & \nilcell{-4}{5} & \nilcell{-4}{5} & \nilcell{-4}{3} & \nilcell{5}{3} \\
 & RU & \nilcell{-9}{2} & \nilcell{-5}{5} & \nilcell{1}{5} & \nilcell{-63}{7} & \nilcell{4}{4} \\
 & TU & \nilcell{-7}{2} & \nilcell{-4}{4} & \nilcell{2}{4} & \nilcell{1}{6} & \nilcell{-2}{3} \\
 & ENCH & \nilcell{-97}{1} & \nilcell{-92}{2} & \nilcell{-7}{5} & \nilcell{-16}{4} & \nilcell{-3}{2} \\
 & ENRU & \nilcell{-96}{2} & \nilcell{-20}{11} & \nilcell{-7}{4} & \nilcell{-84}{4} & \nilcell{-3}{2} \\
 & CHRU & \nilcell{-37}{11} & \nilcell{-93}{3} & \nilcell{-5}{6} & \nilcell{-81}{3} & \nilcell{-1}{2} \\
 & ENCHRU & \nilcell{-96}{2} & \nilcell{-94}{2} & \nilcell{-13}{4} & \nilcell{-85}{2} & \nilcell{-2}{3} \\
\hline
\multirow[t]{10}{*}{\textbf{FT-ENCHTU}} & \textbf{Base} & 96 & 95 & 18 & 14 & 80 \\
 & EN & \nilcell{-92}{5} & \nilcell{-10}{4} & \nilcell{-5}{6} & \nilcell{4}{4} & \nilcell{-49}{11} \\
 & CH & \nilcell{-15}{11} & \nilcell{-82}{2} & \nilcell{2}{7} & \nilcell{1}{3} & \nilcell{-3}{6} \\
 & DE & \nilcell{-24}{4} & \nilcell{-6}{3} & \nilcell{-7}{5} & \nilcell{1}{4} & \nilcell{-6}{7} \\
 & RU & \nilcell{-19}{7} & \nilcell{-7}{3} & \nilcell{-2}{4} & \nilcell{-6}{2} & \nilcell{-4}{5} \\
 & TU & \nilcell{-42}{14} & \nilcell{-3}{4} & \nilcell{0}{6} & \nilcell{0}{5} & \nilcell{-79}{1} \\
 & ENCH & \nilcell{-93}{0} & \nilcell{-89}{4} & \nilcell{-9}{4} & \nilcell{-1}{4} & \nilcell{-56}{12} \\
 & ENTU & \nilcell{-91}{1} & \nilcell{-21}{7} & \nilcell{-9}{1} & \nilcell{-3}{4} & \nilcell{-78}{2} \\
 & CHTU & \nilcell{-67}{8} & \nilcell{-87}{4} & \nilcell{-8}{5} & \nilcell{-2}{4} & \nilcell{-79}{1} \\
 & ENCHTU & \nilcell{-92}{2} & \nilcell{-89}{5} & \nilcell{-10}{3} & \nilcell{-4}{2} & \nilcell{-77}{3} \\
\hline
\multirow[t]{10}{*}{\textbf{FT-ENDERU}} & \textbf{Base} & 95 & 25 & 92 & 94 & 8 \\
 & EN & \nilcell{-91}{4} & \nilcell{-6}{2} & \nilcell{-47}{18} & \nilcell{-10}{7} & \nilcell{-2}{3} \\
 & CH & \nilcell{-8}{3} & \nilcell{-10}{4} & \nilcell{-6}{2} & \nilcell{-5}{5} & \nilcell{2}{4} \\
 & DE & \nilcell{-38}{16} & \nilcell{-8}{4} & \nilcell{-81}{5} & \nilcell{-17}{9} & \nilcell{-1}{3} \\
 & RU & \nilcell{-14}{7} & \nilcell{-7}{4} & \nilcell{-10}{4} & \nilcell{-77}{6} & \nilcell{2}{4} \\
 & TU & \nilcell{-9}{4} & \nilcell{-6}{3} & \nilcell{-7}{4} & \nilcell{-5}{6} & \nilcell{-3}{2} \\
 & ENDE & \nilcell{-92}{3} & \nilcell{-6}{4} & \nilcell{-90}{1} & \nilcell{-29}{9} & \nilcell{-1}{4} \\
 & ENRU & \nilcell{-92}{4} & \nilcell{-12}{6} & \nilcell{-62}{13} & \nilcell{-91}{3} & \nilcell{-1}{2} \\
 & DERU & \nilcell{-61}{13} & \nilcell{-12}{3} & \nilcell{-88}{1} & \nilcell{-89}{4} & \nilcell{-2}{4} \\
 & ENDERU & \nilcell{-93}{2} & \nilcell{-13}{4} & \nilcell{-91}{2} & \nilcell{-90}{5} & \nilcell{-3}{5} \\
\hline
\multirow[t]{10}{*}{\textbf{FT-ENDETU}} & \textbf{Base} & 93 & 20 & 84 & 9 & 83 \\
 & EN & \nilcell{-90}{2} & \nilcell{-6}{4} & \nilcell{-49}{13} & \nilcell{4}{2} & \nilcell{-53}{6} \\
 & CH & \nilcell{-11}{5} & \nilcell{-7}{5} & \nilcell{1}{3} & \nilcell{3}{3} & \nilcell{-1}{3} \\
 & DE & \nilcell{-47}{16} & \nilcell{-5}{5} & \nilcell{-78}{4} & \nilcell{6}{4} & \nilcell{-31}{15} \\
 & RU & \nilcell{-21}{6} & \nilcell{-6}{3} & \nilcell{-9}{5} & \nilcell{-2}{2} & \nilcell{-5}{3} \\
 & TU & \nilcell{-43}{16} & \nilcell{-9}{4} & \nilcell{-33}{13} & \nilcell{6}{3} & \nilcell{-80}{1} \\
 & ENDE & \nilcell{-91}{1} & \nilcell{-9}{3} & \nilcell{-82}{2} & \nilcell{3}{3} & \nilcell{-67}{5} \\
 & ENTU & \nilcell{-89}{1} & \nilcell{-5}{2} & \nilcell{-62}{8} & \nilcell{3}{4} & \nilcell{-81}{1} \\
 & DETU & \nilcell{-74}{9} & \nilcell{-7}{3} & \nilcell{-80}{2} & \nilcell{4}{5} & \nilcell{-81}{2} \\
 & ENDETU & \nilcell{-90}{3} & \nilcell{-9}{2} & \nilcell{-81}{2} & \nilcell{1}{2} & \nilcell{-82}{1} \\
\hline
\multirow[t]{10}{*}{\textbf{FT-CHDERU}} & \textbf{Base} & 27 & 89 & 93 & 85 & 12 \\
 & EN & \nilcell{-21}{3} & \nilcell{-5}{4} & \nilcell{-37}{19} & \nilcell{-14}{6} & \nilcell{-3}{1} \\
 & CH & \nilcell{-9}{4} & \nilcell{-78}{4} & \nilcell{-14}{11} & \nilcell{-3}{3} & \nilcell{-1}{1} \\
 & DE & \nilcell{-8}{6} & \nilcell{-1}{6} & \nilcell{-81}{2} & \nilcell{-6}{6} & \nilcell{-4}{1} \\
 & RU & \nilcell{-9}{6} & \nilcell{-5}{7} & \nilcell{-18}{7} & \nilcell{-72}{6} & \nilcell{-1}{4} \\
 & TU & \nilcell{-8}{6} & \nilcell{-1}{6} & \nilcell{-16}{9} & \nilcell{3}{6} & \nilcell{-6}{3} \\
 & CHDE & \nilcell{-20}{6} & \nilcell{-83}{4} & \nilcell{-91}{1} & \nilcell{-18}{5} & \nilcell{-5}{1} \\
 & CHRU & \nilcell{-13}{3} & \nilcell{-84}{3} & \nilcell{-29}{11} & \nilcell{-83}{2} & \nilcell{-5}{1} \\
 & DERU & \nilcell{-17}{4} & \nilcell{-18}{10} & \nilcell{-90}{2} & \nilcell{-82}{2} & \nilcell{-6}{2} \\
 & CHDERU & \nilcell{-22}{4} & \nilcell{-85}{5} & \nilcell{-92}{0} & \nilcell{-81}{4} & \nilcell{-6}{3} \\
\hline
\multirow[t]{11}{*}{\textbf{FT-ENCHDERUTU}} & \textbf{Base} & 95 & 100 & 95 & 97 & 92 \\
 & EN & \nilcell{-88}{3} & \nilcell{-8}{5} & \nilcell{-45}{14} & \nilcell{-19}{10} & \nilcell{-49}{10} \\
 & CH & \nilcell{-11}{7} & \nilcell{-90}{3} & \nilcell{-13}{9} & \nilcell{-18}{10} & \nilcell{-13}{5} \\
 & DE & \nilcell{-39}{11} & \nilcell{-8}{3} & \nilcell{-84}{9} & \nilcell{-21}{11} & \nilcell{-28}{10} \\
 & RU & \nilcell{-13}{8} & \nilcell{-7}{4} & \nilcell{-13}{8} & \nilcell{-85}{10} & \nilcell{-11}{4} \\
 & TU & \nilcell{-32}{13} & \nilcell{-5}{3} & \nilcell{-30}{13} & \nilcell{-11}{5} & \nilcell{-87}{3} \\
 & ENCHRU & \nilcell{-94}{2} & \nilcell{-99}{1} & \nilcell{-84}{8} & \nilcell{-97}{0} & \nilcell{-75}{15} \\
 & ENDERU & \nilcell{-94}{1} & \nilcell{-68}{18} & \nilcell{-95}{0} & \nilcell{-97}{0} & \nilcell{-79}{8} \\
 & ENDETU & \nilcell{-94}{1} & \nilcell{-58}{20} & \nilcell{-94}{1} & \nilcell{-63}{17} & \nilcell{-89}{1} \\
 & DECHRU & \nilcell{-84}{11} & \nilcell{-99}{1} & \nilcell{-94}{1} & \nilcell{-96}{1} & \nilcell{-64}{21} \\
 & ENCHDERUTU & \nilcell{-94}{1} & \nilcell{-99}{2} & \nilcell{-95}{1} & \nilcell{-97}{0} & \nilcell{-91}{1} \\
\end{longtable}

\subsection{Gemma-2 Unlearning Transfer Across Languages}
\label{app:GemmaTransferResults}

To assess the generalizability of our findings across model architectures, we replicate the cross-lingual unlearning transfer experiments using the \emph{Gemma} model. The experimental setup differs from the methodology described in Section~\ref{sec:CrossLingualTransferText} only in terms of scale due to computational constraints: we evaluate a reduced subset of language combinations and report results from a single experimental run. Despite the reduced scope, the results presented in Table~\ref{tab:GemmaunlearningTransfer} corroborate the trends observed in the main text. Specifically, we observe the same structural properties of transfer regarding linguistic similarity (shared script and language family) and the asymmetry driven by pretraining coverage, suggesting these are fundamental properties of multilingual unlearning rather than artifacts of a specific model.

\FloatBarrier
\begin{longtable}{@{} l@{\hspace{2pt}}l | c | c | c | c | c @{}}
\caption{
Cross-lingual unlearning transfer across fine-tuning families and query languages for Gemma2-9B.
The \textbf{FT} column indicates the fine-tuning language.
Within each row block, the \textbf{Unlearn} column specifies the language on which
unlearning is performed, while \emph{Base} denotes the fine-tuned model before unlearning.
For each query language, we report NLI score \textbf{Forget}
relative to the change in Base model after unlearning in the corresponding unlearned language.
}
\label{tab:GemmaunlearningTransfer}\\
\toprule
\multicolumn{2}{c|}{} & \multicolumn{5}{c}{\textbf{Query}} \\
\midrule
{\bf FT}& \textbf{Unlearn} & \textbf{EN} & \textbf{CH} & \textbf{DE} & \textbf{RU} & \textbf{TU} \\
 & & \textbf{Forget} & \textbf{Forget} & \textbf{Forget} & \textbf{Forget} & \textbf{Forget} \\
\endfirsthead
\toprule
\multicolumn{2}{c|}{} & \multicolumn{5}{c}{\textbf{Query}} \\
\midrule
& \textbf{Unlearn} & \textbf{EN} & \textbf{CH} & \textbf{DE} & \textbf{RU} & \textbf{TU} \\
& & \textbf{Forget} & \textbf{Forget} & \textbf{Forget} & \textbf{Forget} & \textbf{Forget} \\
\midrule
\endhead
\bottomrule
\endfoot
\hline
\multirow[t]{6}{*}{\textbf{FT-EN}} & \textbf{Base} & 97 & 12 & 86 & 47 & 31 \\
 & EN & -94 & -4 & -73 & -40 & -29 \\
 & CH & -11 & -11 & -33 & -34 & -10 \\
 & DE & -54 & 20 & -49 & -13 & -5 \\
 & RU & -26 & 18 & -33 & -21 & 2 \\
 & TU & -41 & 8 & -46 & -16 & -21 \\
\midrule
\multirow[t]{6}{*}{\textbf{FT-CH}} & \textbf{Base} & 14 & 91 & 16 & 47 & 10 \\
 & EN & -14 & -42 & -13 & -33 & -7 \\
 & CH & -5 & -83 & -4 & -40 & 0 \\
 & DE & -5 & -3 & -8 & -24 & -3 \\
 & RU & 14 & 0 & 9 & -25 & 11 \\
 & TU & -3 & -4 & 9 & -14 & -2 \\
\midrule
\multirow[t]{6}{*}{\textbf{FT-DE}} & \textbf{Base} & 53 & 18 & 95 & 27 & 28 \\
 & EN & -37 & 2 & -79 & -18 & -18 \\
 & CH & -5 & -11 & -20 & -1 & 6 \\
 & DE & -48 & -9 & -95 & -21 & -26 \\
 & RU & 3 & 8 & -16 & 7 & 16 \\
 & TU & -17 & 1 & -41 & 9 & -14 \\
\midrule
\multirow[t]{6}{*}{\textbf{FT-RU}} & \textbf{Base} & 24 & 35 & 62 & 90 & 26 \\
 & EN & -14 & -19 & -48 & -60 & -23 \\
 & CH & 18 & -16 & -17 & -10 & 5 \\
 & DE & 21 & -8 & -19 & -2 & 6 \\
 & RU & -5 & -34 & -48 & -90 & -14 \\
 & TU & 18 & -2 & -19 & -7 & -11 \\
\midrule
\multirow[t]{6}{*}{\textbf{FT-TU}} & \textbf{Base} & 19 & 6 & 18 & 15 & 91 \\
 & EN & -19 & 2 & -18 & -9 & -83 \\
 & CH & -6 & 1 & 20 & -2 & -15 \\
 & DE & -3 & 19 & -10 & -3 & -68 \\
 & RU & 4 & 9 & 30 & 2 & -4 \\
 & TU & -13 & -1 & -16 & -4 & -84 \\
\midrule
\multirow[t]{11}{*}{\textbf{FT-ENCHDERUTU}} & \textbf{Base} & 92 & 90 & 97 & 97 & 82 \\
 & EN & -92 & -11 & -70 & -39 & -76 \\
 & CH & -5 & -87 & -10 & -2 & -3 \\
 & DE & -56 & -10 & -97 & -14 & -38 \\
 & RU & -9 & -5 & -9 & -89 & -3 \\
 & TU & -14 & -7 & -15 & -5 & -74 \\
 & ENCHRU & -92 & -89 & -80 & -94 & -79 \\
 & ENDERU & -92 & -34 & -97 & -92 & -71 \\
 & ENDETU & -92 & -38 & -97 & -52 & -81 \\
 & DECHRU & -84 & -89 & -96 & -89 & -50 \\
 & ENCHDERUTU & -92 & -87 & -93 & -94 & -82 \\
\end{longtable}
\FloatBarrier

\subsection{GA and NPO Unlearning Transfer}
\label{app:crossLingualTransferExtraAlgorithms}
To test whether the cross-lingual transfer patterns observed in the main text are specific to DPO-based unlearning, we repeat the same transfer evaluation using two additional unlearning methods: gradient ascent (GA) and negative
preference optimization (NPO). The setup follows
Section~\ref{sec:CrossLingualTransferText}: for each fine-tuned checkpoint, we unlearn in one language and evaluate the change in NLI-based forget-set score across all query languages. As in the main text, more negative values indicate
stronger unlearning transfer. Due to computational cost, we restrict this analysis to the five single-language fine-tuned settings and do not perform additional shuffled forget-set repetitions.

Tables~\ref{tab:GAForgetDelta} and~\ref{tab:NPOForgetDelta} show that the qualitative transfer patterns are broadly consistent across GA and NPO. Although the absolute magnitudes vary across unlearning methods, unlearning still clearly transfers across languages. The qualitative patterns are broadly consistent with the DPO results: transfer is often stronger when the source and query languages share linguistic properties, such as language family or script,
and when the unlearning source is a higher-coverage language. Moreover, unlearning in a language where the model has relatively weak performance can
still transfer to a language where the model performs well. These results suggest that the main cross-lingual transfer findings are not specific to the DPO objective.

\FloatBarrier
\begin{longtable}{@{} l@{\hspace{2pt}}l | c | c | c | c | c @{}}
\caption{Cross-lingual unlearning transfer across fine-tuning languages and query languages under \textbf{GA}. The \textbf{FT} column indicates the fine-tuning language. Within each row block, the \textbf{Unlearn} column specifies the language on which unlearning is performed, while \emph{Base} denotes the fine-tuned model before unlearning. For each query language, we report NLI score \textbf{Forget} relative to the change in Base model after unlearning.}
\label{tab:GAForgetDelta}\\
\toprule
\multicolumn{2}{c|}{} & \multicolumn{5}{c}{\textbf{Query}} \\
\midrule
{\bf FT}& \textbf{Unlearn} & \textbf{EN} & \textbf{CH} & \textbf{DE} & \textbf{RU} & \textbf{TU} \\
 & & \textbf{Forget} & \textbf{Forget} & \textbf{Forget} & \textbf{Forget} & \textbf{Forget} \\
\endfirsthead
\toprule
\multicolumn{2}{c|}{} & \multicolumn{5}{c}{\textbf{Query}} \\
\midrule
& \textbf{Unlearn} & \textbf{EN} & \textbf{CH} & \textbf{DE} & \textbf{RU} & \textbf{TU} \\
 & & \textbf{Forget} & \textbf{Forget} & \textbf{Forget} & \textbf{Forget} & \textbf{Forget} \\
\midrule
\endhead
\bottomrule
\endfoot
\hline
\multirow[t]{6}{*}{\textbf{FT-EN}} & \textbf{Base} & 97 & 12 & 86 & 47 & 31 \\
 & EN & -92 & -6 & -68 & -32 & -22 \\
 & CH & -35 & 2 & -40 & -18 & -8 \\
 & DE & -61 & 14 & -48 & -10 & 0 \\
 & RU & -68 & 15 & -55 & -22 & -13 \\
 & TU & -46 & 6 & -46 & -15 & -15 \\
\midrule
\multirow[t]{6}{*}{\textbf{FT-CH}} & \textbf{Base} & 14 & 91 & 16 & 47 & 10 \\
 & EN & -5 & -79 & -6 & -41 & -5 \\
 & CH & -5 & -40 & -5 & -10 & 20 \\
 & DE & -1 & -37 & 0 & -31 & 1 \\
 & RU & 2 & -54 & 3 & -34 & 0 \\
 & TU & 8 & -41 & 9 & -19 & 4 \\
\midrule
\multirow[t]{6}{*}{\textbf{FT-DE}} & \textbf{Base} & 53 & 18 & 95 & 27 & 28 \\
 & EN & -32 & -3 & -85 & -5 & -21 \\
 & CH & -7 & -14 & -56 & -3 & -3 \\
 & DE & -39 & -11 & -89 & -16 & -27 \\
 & RU & -16 & 11 & -59 & 4 & -10 \\
 & TU & -19 & 4 & -55 & 5 & -8 \\
\midrule
\multirow[t]{6}{*}{\textbf{FT-RU}} & \textbf{Base} & 24 & 35 & 62 & 90 & 26 \\
 & EN & -5 & -19 & -52 & -72 & -18 \\
 & CH & 5 & -19 & -30 & -45 & -11 \\
 & DE & 10 & -5 & -36 & -50 & -6 \\
 & RU & -12 & -28 & -42 & -79 & -24 \\
 & TU & 10 & 0 & -31 & -38 & -10 \\
\midrule
\multirow[t]{6}{*}{\textbf{FT-TU}} & \textbf{Base} & 19 & 6 & 18 & 15 & 91 \\
 & EN & -3 & 8 & -9 & -1 & -79 \\
 & CH & 25 & 0 & 16 & 0 & -42 \\
 & DE & 0 & 12 & -7 & 3 & -74 \\
 & RU & 8 & 13 & 8 & -2 & -64 \\
 & TU & -3 & 24 & 8 & 29 & -69 \\
\end{longtable}
\FloatBarrier

\FloatBarrier
\begin{longtable}{@{} l@{\hspace{2pt}}l | c | c | c | c | c @{}}
\caption{Cross-lingual unlearning transfer across fine-tuning languages and query languages under \textbf{NPO}. The \textbf{FT} column indicates the fine-tuning language. Within each row block, the \textbf{Unlearn} column specifies the language on which unlearning is performed, while \emph{Base} denotes the fine-tuned model before unlearning. For each query language, we report NLI score \textbf{Forget} relative to the change in Base model after unlearning.}
\label{tab:NPOForgetDelta}\\
\toprule
\multicolumn{2}{c|}{} & \multicolumn{5}{c}{\textbf{Query}} \\
\midrule
{\bf FT}& \textbf{Unlearn} & \textbf{EN} & \textbf{CH} & \textbf{DE} & \textbf{RU} & \textbf{TU} \\
 & & \textbf{Forget} & \textbf{Forget} & \textbf{Forget} & \textbf{Forget} & \textbf{Forget} \\
\endfirsthead
\toprule
\multicolumn{2}{c|}{} & \multicolumn{5}{c}{\textbf{Query}} \\
\midrule
& \textbf{Unlearn} & \textbf{EN} & \textbf{CH} & \textbf{DE} & \textbf{RU} & \textbf{TU} \\
 & & \textbf{Forget} & \textbf{Forget} & \textbf{Forget} & \textbf{Forget} & \textbf{Forget} \\
\midrule
\endhead
\bottomrule
\endfoot
\hline
\multirow[t]{6}{*}{\textbf{FT-EN}} & \textbf{Base} & 97 & 12 & 86 & 47 & 31 \\
 & EN & -85 & -6 & -73 & -34 & -19 \\
 & CH & -28 & -4 & -30 & -22 & -10 \\
 & DE & -35 & 18 & -50 & -17 & 0 \\
 & RU & -35 & 10 & -36 & -28 & 3 \\
 & TU & -46 & 11 & -38 & -21 & -21 \\
\midrule
\multirow[t]{6}{*}{\textbf{FT-CH}} & \textbf{Base} & 14 & 91 & 16 & 47 & 10 \\
 & EN & -10 & -54 & -6 & -38 & -2 \\
 & CH & -6 & -69 & -3 & -24 & 1 \\
 & DE & -3 & -11 & -5 & -32 & 1 \\
 & RU & 24 & 2 & 13 & -15 & 6 \\
 & TU & 13 & -4 & 8 & -15 & 2 \\
\midrule
\multirow[t]{6}{*}{\textbf{FT-DE}} & \textbf{Base} & 53 & 18 & 95 & 27 & 28 \\
 & EN & -22 & 10 & -54 & 2 & -1 \\
 & CH & -6 & -12 & -35 & -10 & 4 \\
 & DE & -41 & -12 & -89 & -14 & -22 \\
 & RU & 12 & 17 & -30 & 3 & 14 \\
 & TU & 1 & 15 & -29 & 13 & -3 \\
\midrule
\multirow[t]{6}{*}{\textbf{FT-RU}} & \textbf{Base} & 24 & 35 & 62 & 90 & 26 \\
 & EN & -12 & -15 & -51 & -56 & -15 \\
 & CH & 18 & -16 & -9 & -6 & 5 \\
 & DE & 30 & 9 & -12 & -6 & 18 \\
 & RU & -7 & -30 & -42 & -81 & -17 \\
 & TU & 23 & -1 & -12 & -4 & 4 \\
\midrule
\multirow[t]{6}{*}{\textbf{FT-TU}} & \textbf{Base} & 19 & 6 & 18 & 15 & 91 \\
 & EN & -14 & 1 & -5 & -5 & -60 \\
 & CH & 10 & 0 & 28 & 2 & -20 \\
 & DE & -6 & 15 & -13 & 15 & -26 \\
 & RU & 1 & 21 & 24 & -10 & -12 \\
 & TU & -5 & 14 & 7 & 10 & -77 \\
\end{longtable}
\FloatBarrier

\subsection{Retain Set Results}
\label{app:retainSetResults}

For completeness, we report the retain set results omitted from the main text. The \emph{Base} row gives the retain performance of the fine-tuned model before unlearning, while the remaining rows report the change in retain performance after unlearning relative to the corresponding Base model. The retain results show little evidence of systematic cross-lingual transfer. Although retain performance can decrease when the query language is the same as the unlearned language, the off-diagonal changes across different query languages are generally small and do not form consistent patterns across fine-tuning families, language families, or scripts. This contrasts with the forget-set results, where cross-lingual transfer is substantially larger and more structured. These results support our focus on forget-set transfer in the main text: the main cross-lingual effect of unlearning appears in forgetting behavior, while retain-set transfer across languages is comparatively limited.

\FloatBarrier
\begin{longtable}{@{} l@{\hspace{2pt}}l | c | c | c | c | c @{}}
\caption{Retain-set performance under \textbf{DPO} for Qwen2.5-7B. The \textbf{FT} column indicates the fine-tuning language, and \textbf{Unlearn} specifies the unlearned language. The \emph{Base} row reports absolute Retain NLI score. DPO rows report $\Delta$Retain relative to the corresponding Base model; values are means $\pm$ 95\% confidence intervals over the selected shuffles.}
\label{tab:dpo-retain-delta}\\
\toprule
\multicolumn{2}{c|}{} & \multicolumn{5}{c}{\textbf{Query}} \\
\midrule
{\bf FT}& \textbf{Unlearn} & \textbf{EN} & \textbf{CH} & \textbf{DE} & \textbf{RU} & \textbf{TU} \\
 & & \textbf{Retain} & \textbf{Retain} & \textbf{Retain} & \textbf{Retain} & \textbf{Retain} \\
\endfirsthead
\toprule
\multicolumn{2}{c|}{} & \multicolumn{5}{c}{\textbf{Query}} \\
\midrule
& \textbf{Unlearn} & \textbf{EN} & \textbf{CH} & \textbf{DE} & \textbf{RU} & \textbf{TU} \\
 & & \textbf{Retain} & \textbf{Retain} & \textbf{Retain} & \textbf{Retain} & \textbf{Retain} \\
\midrule
\endhead
\bottomrule
\endfoot
\hline
\multirow[t]{6}{*}{\textbf{FT-EN}} & \textbf{Base} & 92 & 9 & 17 & 12 & 8 \\
 & EN & \nilcell{-7}{2} & \nilcell{0}{0} & \nilcell{-1}{0} & \nilcell{-1}{0} & \nilcell{-1}{0} \\
 & CH & \nilcell{-1}{0} & \nilcell{-1}{0} & \nilcell{0}{0} & \nilcell{0}{0} & \nilcell{0}{0} \\
 & DE & \nilcell{-1}{0} & \nilcell{0}{0} & \nilcell{2}{2} & \nilcell{0}{1} & \nilcell{0}{1} \\
 & RU & \nilcell{-1}{0} & \nilcell{-1}{0} & \nilcell{-1}{0} & \nilcell{-2}{0} & \nilcell{0}{0} \\
 & TU & \nilcell{-1}{0} & \nilcell{0}{0} & \nilcell{0}{1} & \nilcell{0}{0} & \nilcell{-2}{0} \\
\midrule
\multirow[t]{6}{*}{\textbf{FT-CH}} & \textbf{Base} & 8 & 92 & 8 & 10 & 6 \\
 & EN & \nilcell{-2}{0} & \nilcell{-1}{0} & \nilcell{-1}{0} & \nilcell{-1}{0} & \nilcell{0}{0} \\
 & CH & \nilcell{-1}{0} & \nilcell{-6}{1} & \nilcell{-1}{0} & \nilcell{-2}{0} & \nilcell{0}{0} \\
 & DE & \nilcell{0}{0} & \nilcell{-1}{0} & \nilcell{-1}{0} & \nilcell{0}{0} & \nilcell{0}{0} \\
 & RU & \nilcell{0}{0} & \nilcell{-1}{0} & \nilcell{-1}{0} & \nilcell{-2}{1} & \nilcell{0}{0} \\
 & TU & \nilcell{0}{0} & \nilcell{-1}{0} & \nilcell{0}{0} & \nilcell{0}{0} & \nilcell{-1}{0} \\
\midrule
\multirow[t]{6}{*}{\textbf{FT-DE}} & \textbf{Base} & 14 & 9 & 92 & 10 & 7 \\
 & EN & \nilcell{-4}{1} & \nilcell{0}{0} & \nilcell{-1}{0} & \nilcell{0}{0} & \nilcell{0}{0} \\
 & CH & \nilcell{0}{0} & \nilcell{-1}{1} & \nilcell{-1}{0} & \nilcell{0}{0} & \nilcell{0}{0} \\
 & DE & \nilcell{0}{0} & \nilcell{0}{0} & \nilcell{-8}{1} & \nilcell{0}{0} & \nilcell{-1}{0} \\
 & RU & \nilcell{0}{0} & \nilcell{0}{0} & \nilcell{-1}{0} & \nilcell{-2}{1} & \nilcell{0}{0} \\
 & TU & \nilcell{0}{0} & \nilcell{0}{0} & \nilcell{-1}{0} & \nilcell{0}{0} & \nilcell{-1}{0} \\
\midrule
\multirow[t]{6}{*}{\textbf{FT-RU}} & \textbf{Base} & 10 & 9 & 9 & 93 & 6 \\
 & EN & \nilcell{-3}{1} & \nilcell{0}{0} & \nilcell{-1}{0} & \nilcell{-1}{0} & \nilcell{-1}{0} \\
 & CH & \nilcell{0}{0} & \nilcell{-1}{0} & \nilcell{0}{0} & \nilcell{-1}{0} & \nilcell{0}{0} \\
 & DE & \nilcell{0}{0} & \nilcell{0}{0} & \nilcell{-1}{0} & \nilcell{-1}{0} & \nilcell{0}{0} \\
 & RU & \nilcell{0}{0} & \nilcell{0}{0} & \nilcell{-1}{0} & \nilcell{-10}{2} & \nilcell{0}{0} \\
 & TU & \nilcell{0}{0} & \nilcell{0}{0} & \nilcell{0}{0} & \nilcell{-1}{0} & \nilcell{-1}{0} \\
\midrule
\multirow[t]{6}{*}{\textbf{FT-TU}} & \textbf{Base} & 11 & 9 & 10 & 10 & 93 \\
 & EN & \nilcell{-7}{1} & \nilcell{-1}{0} & \nilcell{-1}{0} & \nilcell{-2}{0} & \nilcell{-2}{0} \\
 & CH & \nilcell{-2}{0} & \nilcell{-3}{1} & \nilcell{0}{0} & \nilcell{-2}{0} & \nilcell{-2}{0} \\
 & DE & \nilcell{-1}{0} & \nilcell{0}{0} & \nilcell{-2}{0} & \nilcell{-2}{0} & \nilcell{-2}{0} \\
 & RU & \nilcell{-2}{0} & \nilcell{-1}{0} & \nilcell{-1}{0} & \nilcell{-6}{1} & \nilcell{-2}{0} \\
 & TU & \nilcell{-1}{0} & \nilcell{0}{0} & \nilcell{0}{0} & \nilcell{-1}{0} & \nilcell{-11}{1} \\
\midrule
\multirow[t]{10}{*}{\textbf{FT-ENCHRU}} & \textbf{Base} & 93 & 93 & 26 & 90 & 13 \\
 & EN & \nilcell{-4}{1} & \nilcell{-1}{0} & \nilcell{-2}{0} & \nilcell{-1}{0} & \nilcell{-1}{0} \\
 & CH & \nilcell{-1}{0} & \nilcell{-3}{1} & \nilcell{-2}{1} & \nilcell{-1}{0} & \nilcell{-1}{0} \\
 & DE & \nilcell{0}{0} & \nilcell{0}{0} & \nilcell{-1}{0} & \nilcell{0}{0} & \nilcell{-1}{0} \\
 & RU & \nilcell{-1}{0} & \nilcell{-2}{0} & \nilcell{-2}{0} & \nilcell{-5}{1} & \nilcell{-2}{0} \\
 & TU & \nilcell{0}{0} & \nilcell{0}{0} & \nilcell{0}{0} & \nilcell{0}{0} & \nilcell{-1}{0} \\
 & ENCH & \nilcell{-3}{1} & \nilcell{-4}{1} & \nilcell{-2}{0} & \nilcell{-1}{0} & \nilcell{-2}{0} \\
 & ENRU & \nilcell{-3}{1} & \nilcell{-2}{1} & \nilcell{-3}{1} & \nilcell{-5}{1} & \nilcell{-2}{0} \\
 & CHRU & \nilcell{-2}{0} & \nilcell{-4}{1} & \nilcell{-3}{1} & \nilcell{-6}{1} & \nilcell{-2}{0} \\
 & ENCHRU & \nilcell{-1}{0} & \nilcell{-3}{1} & \nilcell{-1}{0} & \nilcell{-2}{0} & \nilcell{-1}{0} \\
\midrule
\multirow[t]{10}{*}{\textbf{FT-ENCHTU}} & \textbf{Base} & 92 & 91 & 23 & 20 & 82 \\
 & EN & \nilcell{-5}{1} & \nilcell{-1}{0} & \nilcell{-2}{0} & \nilcell{-3}{0} & \nilcell{-1}{0} \\
 & CH & \nilcell{-1}{0} & \nilcell{-4}{1} & \nilcell{-2}{0} & \nilcell{-3}{0} & \nilcell{-1}{0} \\
 & DE & \nilcell{0}{0} & \nilcell{0}{0} & \nilcell{-1}{0} & \nilcell{-1}{0} & \nilcell{0}{0} \\
 & RU & \nilcell{0}{0} & \nilcell{0}{0} & \nilcell{-1}{0} & \nilcell{-2}{1} & \nilcell{0}{0} \\
 & TU & \nilcell{-1}{0} & \nilcell{-1}{0} & \nilcell{-2}{0} & \nilcell{-3}{0} & \nilcell{-8}{1} \\
 & ENCH & \nilcell{-3}{1} & \nilcell{-3}{1} & \nilcell{-2}{0} & \nilcell{-2}{0} & \nilcell{-2}{1} \\
 & ENTU & \nilcell{-2}{1} & \nilcell{-1}{0} & \nilcell{-2}{0} & \nilcell{-3}{0} & \nilcell{-4}{0} \\
 & CHTU & \nilcell{-1}{0} & \nilcell{-3}{1} & \nilcell{-2}{0} & \nilcell{-3}{0} & \nilcell{-4}{1} \\
 & ENCHTU & \nilcell{-2}{1} & \nilcell{-2}{1} & \nilcell{0}{0} & \nilcell{-1}{0} & \nilcell{-2}{1} \\
\midrule
\multirow[t]{10}{*}{\textbf{FT-ENDERU}} & \textbf{Base} & 94 & 18 & 89 & 90 & 12 \\
 & EN & \nilcell{-6}{1} & \nilcell{-1}{0} & \nilcell{-2}{0} & \nilcell{0}{0} & \nilcell{-2}{0} \\
 & CH & \nilcell{-1}{0} & \nilcell{0}{0} & \nilcell{0}{0} & \nilcell{0}{0} & \nilcell{-1}{0} \\
 & DE & \nilcell{-3}{0} & \nilcell{-1}{0} & \nilcell{-6}{1} & \nilcell{-1}{0} & \nilcell{-2}{0} \\
 & RU & \nilcell{-2}{1} & \nilcell{-1}{0} & \nilcell{-2}{1} & \nilcell{-5}{2} & \nilcell{-2}{0} \\
 & TU & \nilcell{0}{0} & \nilcell{0}{0} & \nilcell{0}{0} & \nilcell{0}{0} & \nilcell{-1}{0} \\
 & ENDE & \nilcell{-3}{1} & \nilcell{-1}{0} & \nilcell{-3}{0} & \nilcell{-1}{0} & \nilcell{-2}{0} \\
 & ENRU & \nilcell{-3}{1} & \nilcell{-1}{0} & \nilcell{-2}{1} & \nilcell{-4}{1} & \nilcell{-2}{0} \\
 & DERU & \nilcell{-3}{1} & \nilcell{-1}{0} & \nilcell{-4}{0} & \nilcell{-5}{1} & \nilcell{-2}{0} \\
 & ENDERU & \nilcell{-2}{1} & \nilcell{0}{0} & \nilcell{-2}{0} & \nilcell{-2}{0} & \nilcell{-1}{0} \\
\midrule
\multirow[t]{10}{*}{\textbf{FT-ENDETU}} & \textbf{Base} & 92 & 16 & 91 & 20 & 84 \\
 & EN & \nilcell{-6}{1} & \nilcell{-2}{0} & \nilcell{-2}{0} & \nilcell{-2}{1} & \nilcell{-2}{1} \\
 & CH & \nilcell{0}{0} & \nilcell{-1}{0} & \nilcell{0}{0} & \nilcell{-1}{0} & \nilcell{0}{0} \\
 & DE & \nilcell{-3}{1} & \nilcell{-2}{0} & \nilcell{-8}{1} & \nilcell{-3}{0} & \nilcell{-3}{1} \\
 & RU & \nilcell{0}{0} & \nilcell{-1}{0} & \nilcell{-1}{0} & \nilcell{-1}{0} & \nilcell{0}{0} \\
 & TU & \nilcell{-1}{0} & \nilcell{-2}{0} & \nilcell{-2}{0} & \nilcell{-2}{0} & \nilcell{-8}{0} \\
 & ENDE & \nilcell{-2}{1} & \nilcell{-2}{0} & \nilcell{-4}{1} & \nilcell{-2}{0} & \nilcell{-2}{1} \\
 & ENTU & \nilcell{-2}{1} & \nilcell{-2}{0} & \nilcell{-2}{1} & \nilcell{-2}{0} & \nilcell{-5}{0} \\
 & DETU & \nilcell{-2}{0} & \nilcell{-2}{0} & \nilcell{-4}{0} & \nilcell{-3}{0} & \nilcell{-5}{0} \\
 & ENDETU & \nilcell{-1}{0} & \nilcell{-1}{0} & \nilcell{-2}{0} & \nilcell{-1}{0} & \nilcell{-3}{0} \\
\midrule
\multirow[t]{10}{*}{\textbf{FT-CHDERU}} & \textbf{Base} & 25 & 91 & 88 & 89 & 10 \\
 & EN & \nilcell{0}{0} & \nilcell{0}{0} & \nilcell{-1}{0} & \nilcell{0}{0} & \nilcell{0}{0} \\
 & CH & \nilcell{-2}{1} & \nilcell{-3}{1} & \nilcell{-1}{0} & \nilcell{-1}{0} & \nilcell{-1}{0} \\
 & DE & \nilcell{-2}{0} & \nilcell{-1}{0} & \nilcell{-5}{1} & \nilcell{-2}{0} & \nilcell{-1}{0} \\
 & RU & \nilcell{-2}{0} & \nilcell{-1}{0} & \nilcell{-2}{0} & \nilcell{-6}{1} & \nilcell{-1}{0} \\
 & TU & \nilcell{0}{0} & \nilcell{0}{0} & \nilcell{-1}{0} & \nilcell{0}{0} & \nilcell{-1}{0} \\
 & CHDE & \nilcell{-3}{1} & \nilcell{-4}{1} & \nilcell{-5}{1} & \nilcell{-2}{0} & \nilcell{-1}{0} \\
 & CHRU & \nilcell{-3}{0} & \nilcell{-4}{1} & \nilcell{-2}{0} & \nilcell{-6}{1} & \nilcell{-2}{0} \\
 & DERU & \nilcell{-3}{0} & \nilcell{-2}{0} & \nilcell{-5}{1} & \nilcell{-6}{1} & \nilcell{-2}{0} \\
 & CHDERU & \nilcell{-1}{1} & \nilcell{-2}{1} & \nilcell{-3}{1} & \nilcell{-3}{1} & \nilcell{-1}{0} \\
\midrule
\multirow[t]{11}{*}{\textbf{FT-ENCHDERUTU}} & \textbf{Base} & 96 & 96 & 93 & 94 & 88 \\
 & EN & \nilcell{-3}{1} & \nilcell{0}{0} & \nilcell{-1}{0} & \nilcell{-1}{0} & \nilcell{-1}{0} \\
 & CH & \nilcell{-1}{0} & \nilcell{-3}{1} & \nilcell{-1}{0} & \nilcell{-1}{0} & \nilcell{-1}{0} \\
 & DE & \nilcell{-1}{0} & \nilcell{-1}{0} & \nilcell{-4}{1} & \nilcell{-1}{0} & \nilcell{-2}{0} \\
 & RU & \nilcell{-1}{0} & \nilcell{-1}{0} & \nilcell{-1}{0} & \nilcell{-4}{1} & \nilcell{-1}{0} \\
 & TU & \nilcell{-1}{0} & \nilcell{-1}{0} & \nilcell{-1}{0} & \nilcell{-1}{0} & \nilcell{-5}{1} \\
 & ENCHRU & \nilcell{-1}{0} & \nilcell{-2}{0} & \nilcell{-1}{0} & \nilcell{-2}{0} & \nilcell{-1}{0} \\
 & ENDERU & \nilcell{-1}{0} & \nilcell{-1}{0} & \nilcell{-2}{0} & \nilcell{-2}{0} & \nilcell{-1}{0} \\
 & ENDETU & \nilcell{-1}{1} & \nilcell{-1}{0} & \nilcell{-2}{0} & \nilcell{-1}{0} & \nilcell{-2}{0} \\
 & DECHRU & \nilcell{-1}{0} & \nilcell{-2}{0} & \nilcell{-2}{1} & \nilcell{-2}{0} & \nilcell{-1}{0} \\
 & ENCHDERUTU & \nilcell{-1}{0} & \nilcell{-1}{0} & \nilcell{-1}{0} & \nilcell{-1}{0} & \nilcell{-1}{0} \\
\end{longtable}
\FloatBarrier

\section{Prompting}
\label{app:Prompting}
We enforce the output language at inference time by appending a short, language-specific
instruction to the user query. Given input question $x$ in language $q$ and fine-tuned language $\ell$, we form $x' = x + I_\ell$. For each question, we append the
corresponding instruction as follows:
\begin{itemize}
    \item \textbf{English}: \emph{Question} + ``You must only answer this question in English.''
    
    \item \textbf{Chinese}: \emph{Question} + ``\begin{CJK*}{UTF8}{gbsn}你必须仅用中文回答这个问题\end{CJK*}''
    
    \item \textbf{German}: \emph{Question} + ``Bitte beantworte diese Frage ausschließlich auf Deutsch.''
    
    \item \textbf{Russian}: \emph{Question} + ``\foreignlanguage{russian}{Пожалуйста, отвечай на этот вопрос только на русском языке.}''
    
    \item \textbf{Turkish}: \emph{Question} + ``Lütfen bu soruyu yalnızca Türkçe olarak cevapla.''
\end{itemize}

\section{Hidden Representation Analysis}
\label{app:hiddenRepresentationAnalysis}

\subsection{Cosine Similarity for Chinese}
\label{app:cosChinese}

\begin{figure*}[t]
    \centering
    \makebox[\textwidth][c]{%
        \includegraphics[width=1\textwidth,height=20\textheight,keepaspectratio]{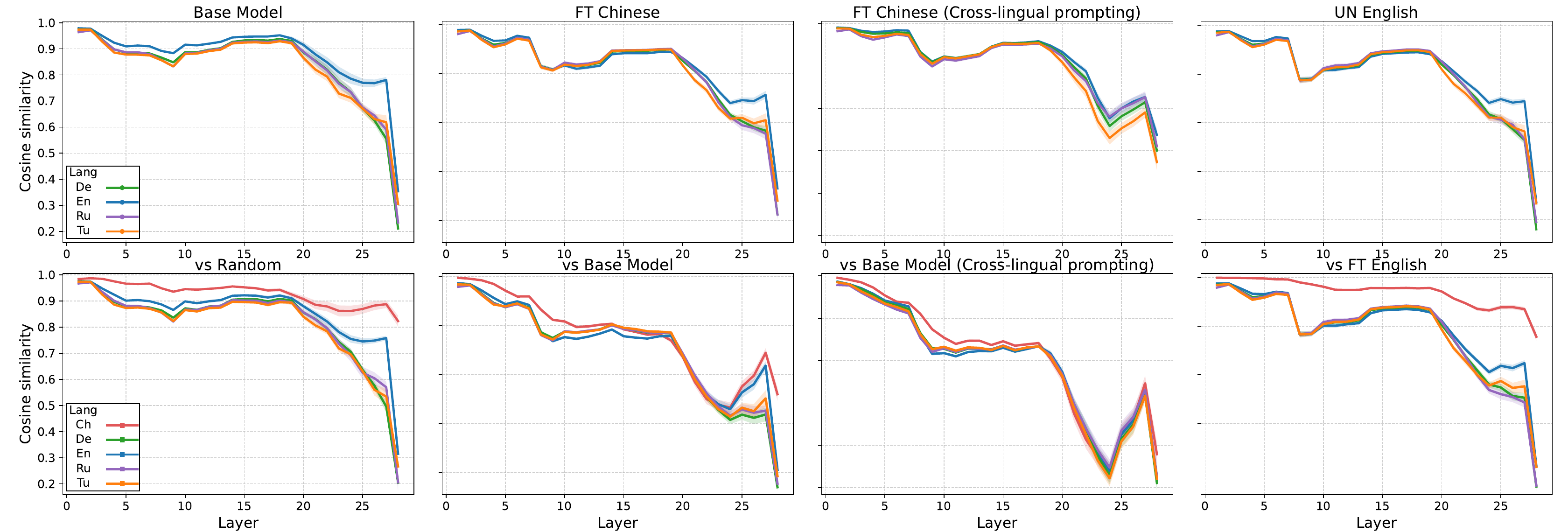}
    }
    \caption{
    Layer-wise cosine similarity between the hidden state ${h}^{(l)}_{m}(x_\ell)$ of the final input token for Chinese, English, German, Russian, and Turkish questions, across three variants of Qwen2.5-7B (see panel titles for settings).
    \textit{FT} denotes fine-tuning, and \textit{UN} denotes unlearning.
    \textit{Cross-lingual prompting} means that for non-English questions, the model is instructed to answer in English.
    \textbf{Bottom row:} control comparisons, including English vs.\ randomly paired non-equivalent questions from $\mathcal{D}^{\text{retain}}_{\ell}$ (\emph{vs.\ Random}) and cross-model comparisons using English representations from the corresponding top-row model (\emph{vs\ Base Model}, \emph{vs\ FT English}).
    Curves show the mean across questions, shaded regions denote 95\% confidence intervals.
    }
    \label{fig:ch_combined}
\end{figure*}

To verify that the representational patterns observed in the main text are not specific to English (a high-resource Indo-European language), we replicate the layer-wise cosine similarity analysis using Chinese as the anchor language. In this setup, the model is fine-tuned on Chinese, and we measure the alignment between Chinese representations and those of other languages (English, German, Russian, Turkish) for the same underlying questions. Figure~\ref{fig:ch_combined} presents these results. Consistent with the English-centric analysis.

\subsection{Impact of Unlearning for Fine-tuned Language}
\begin{figure}[h]
    \centering
    \includegraphics[width=0.8\linewidth]{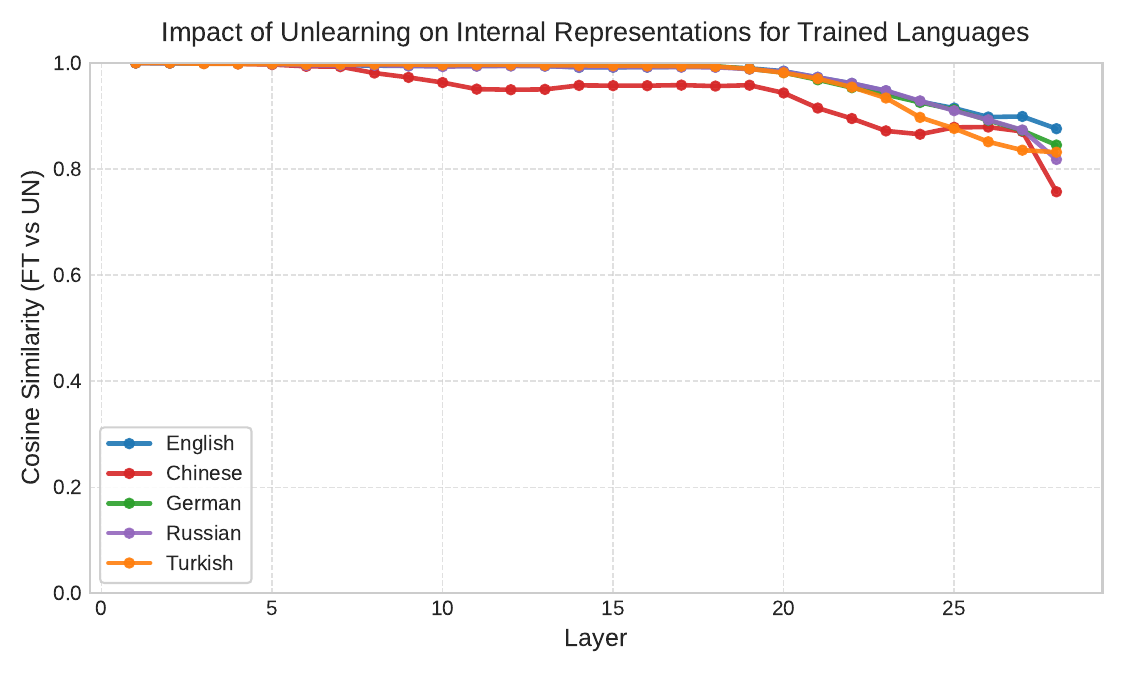}
    \caption{We compare the cosine similarity of hidden representations before and after unlearning within the same language for which the model was finetuned. The results show that the early and mid layers are minimally affected, while the majority of the similarity drop occurs in the later layers.}
    \label{fig:impact_FT_UN}
\end{figure}

To understand the mechanism of unlearning at a granular level, we investigate how the internal representations of the \emph{same} input evolve from the fine-tuned model (\FT) to the unlearned model (\UN). Figure~\ref{fig:impact_FT_UN} plots the layer-wise cosine similarity between $h^{(l)}_{\FT}(x)$ and $h^{(l)}_{\UN}(x)$ for the five languages. The results reveal a distinct ``late-stage'' intervention pattern. For the first two-thirds of the layers, the cosine similarity remains very high, indicating that the unlearning algorithm induces negligible changes to the early semantic processing and feature extraction stages, with the exception of a slightly larger drop observed for Chinese. The divergence begins sharply only in the later layers, where the similarity drops as the model after unlearn pushes the representation away from the original target output. This trend mirrors the impact of unlearning observed in multilingual settings, and this finding provides mechanistic evidence that unlearning functions primarily by altering the decoding trajectory at the output layers, leaving the deep semantic encoding of the sensitive knowledge largely intact.

\label{app: ImpactUnlearningFT}

\subsection{Principal Component and Distance Analysis}
\label{app:PCAdisAnalysis}
\noindent\textbf{Setup.}
Using the same hidden representations $h^{(l)}_{m}(x_\ell)$ in Section~\ref{sec:cosineSimilarity}, we study how unlearning reshapes the geometry of the forget representations. We first $\mathrm{L}_2$-normalize the representation,
$
  \tilde{h}^{(l)}_{m}(x_\ell)
  = \frac{h^{(l)}_{m}(x_\ell)}
         {\bigl\|h^{(l)}_{m}(x_\ell)\bigr\|_2}
$
and collect the normalized vectors for all questions in the forget set:
$
  S^{(l)}_{m}(\ell)
  = \bigl\{
      \tilde{h}^{(l)}_{m}(x_\ell)
      : x_\ell \in \mathcal{D}^{\text{forget}}_\ell
    \bigr\},
   m \in \{\Base,\FT,\UN\}
$.
We define the centroid for a given layer as
$
  c^{(l)}_{m}(\ell)
  = \frac{1}{\bigl|S^{(l)}_{m}(\ell)\bigr|}
    \sum_{h \in S^{(l)}_{m}(\ell)} h
$.
We then quantify representation changes in $S^{(l)}_{m}(\ell)$ using two complementary distance measures.
First, \emph{centroid distance} captures \emph{global} distributional shifts by computing the $\mathrm{L}_2$ distance between centroids of two model variants (e.g., $\bigl\|c^{(l)}_{\FT}(\ell) - c^{(l)}_{\UN}(\ell)\bigr\|_2$).
Second, to capture the \emph{per-example} effect of unlearning, we compute the pairwise distance
$d^{(l)}(x_\ell) := \bigl\|\tilde{h}^{(l)}_{m_1}(x_\ell) - \tilde{h}^{(l)}_{m_2}(x_\ell)\bigr\|_2$
for each forget question $x_\ell$ at layer $l$, for $m_1,m_2 \in \{\Base,\FT,\UN\}$, and average over all forget questions to obtain \emph{average pairwise distances}.

\begin{figure*}[h]
    \centering
    \includegraphics[
      width=\linewidth,
    ]{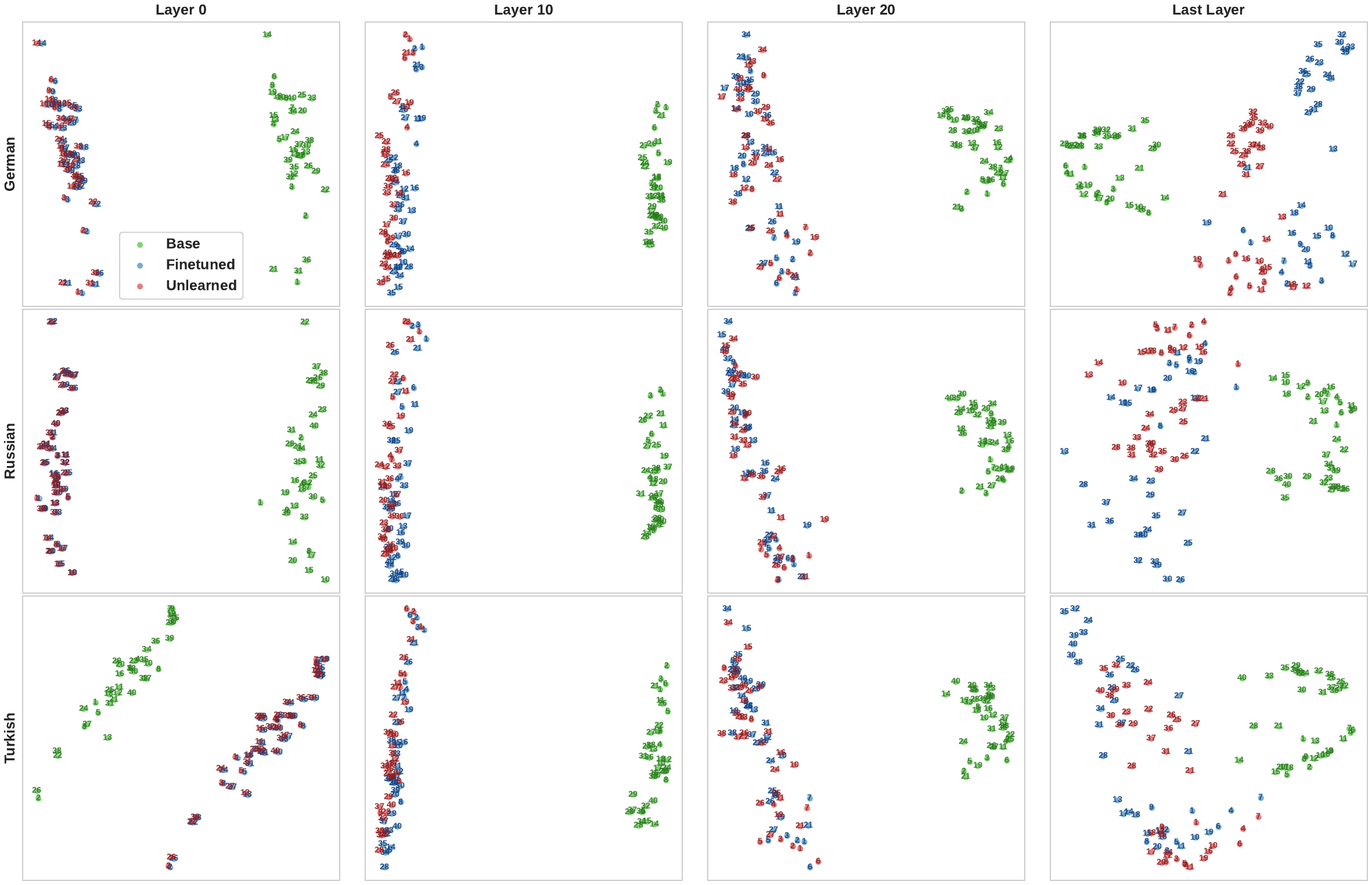}
    \caption{PCA separation across layers for German (top), Russian (middle) and Turkish (bottom) questions. Numbers indicate question indices, with identical numbers referring to the same questions across different testing model variants.}
    \label{fig:PCADeRuTu}
\end{figure*}

\begin{figure*}[t]
    \centering
    \begin{subfigure}[t]{0.48\linewidth}
        \centering
        \includegraphics[width=\linewidth]{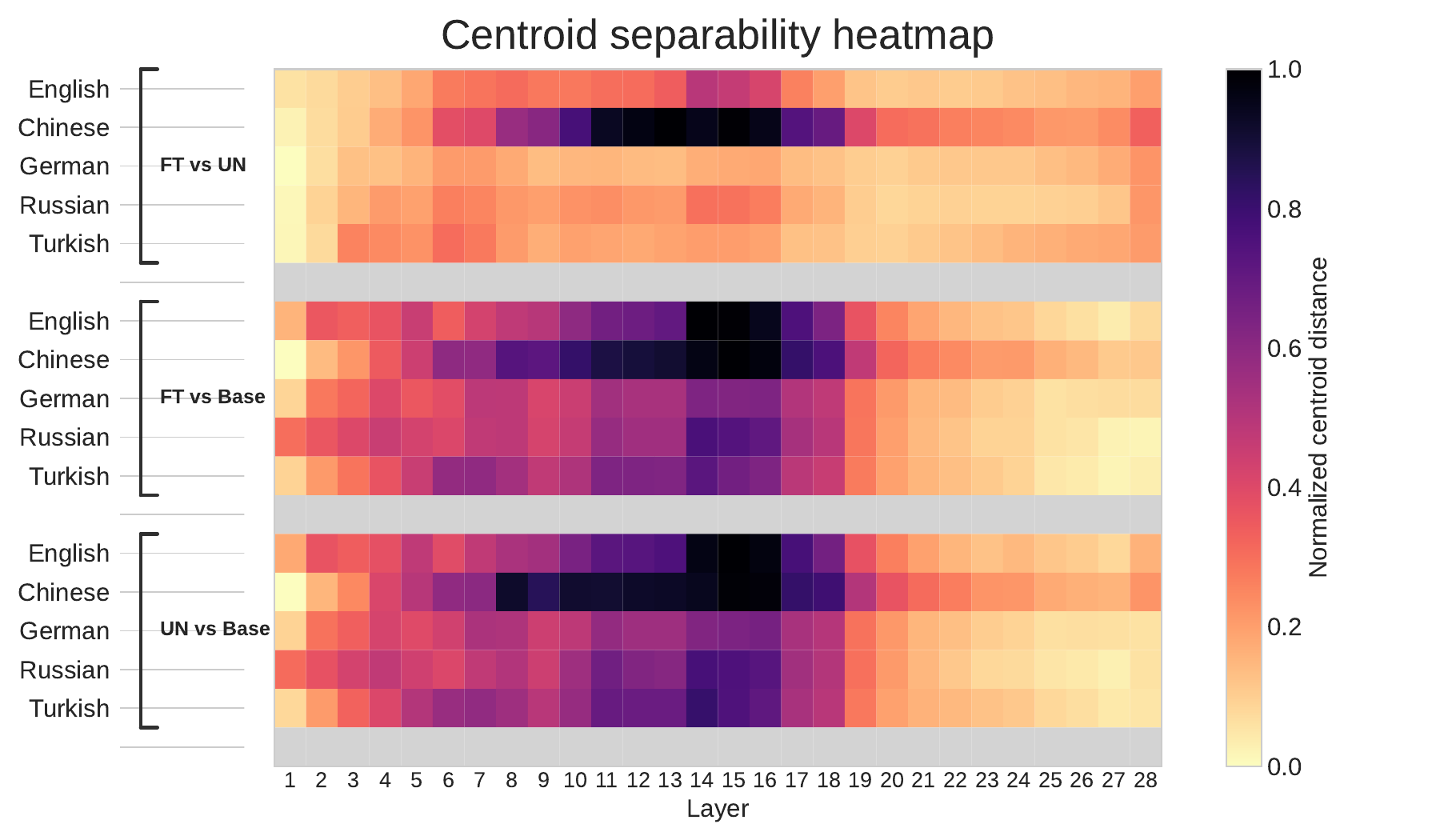}
        \caption{Centroid Distance for three group}
        \label{fig:centroid_heatmap}
    \end{subfigure}
    \hfill
    \begin{subfigure}[t]{0.48\linewidth}
        \centering
        \includegraphics[width=\linewidth]{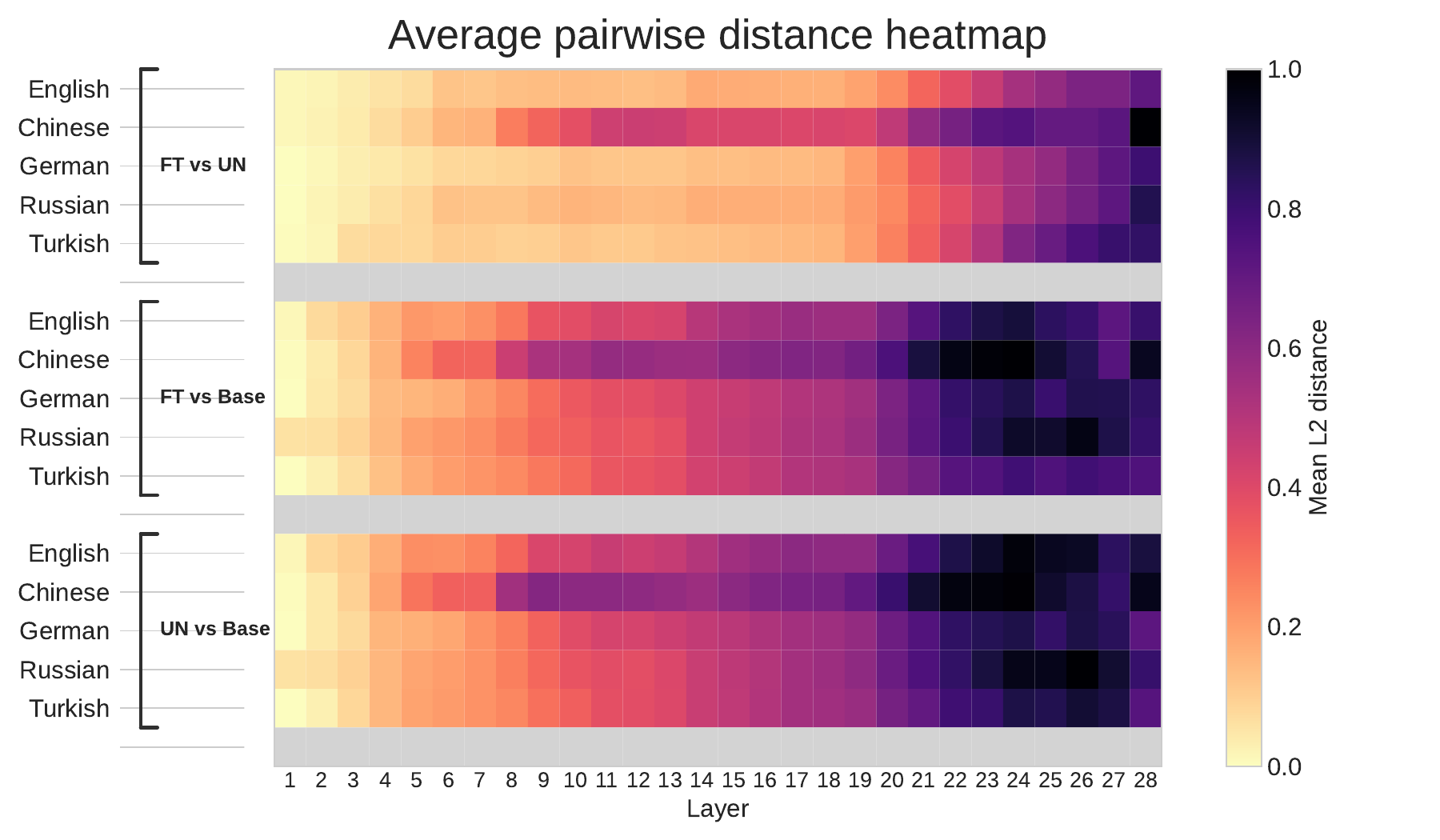}
        \caption{Average Pair-Wise Distance for three groups}
        \label{fig:distance_heatmap}
    \end{subfigure}
    \caption{The left heatmap capture the impact of unlearning for global geometry, while the right heatmap capture per-sample effect of unlearning.}
    \label{fig:heatmap}
\end{figure*}

\noindent\textbf{Results.}
The average pairwise distance $|h_{\text{FT}}^{(l)}(i)-h_{\text{UN}}^{(l)}(i)|$ mirrors the cosine similarity, remaining small at early and middle layers but growing sharply near the output (Figure~\ref{fig:distance_heatmap}). In contrast, the distance between \FT{} and \UN{} centroids remains relatively constant across all layers for most languages (Figure~\ref{fig:centroid_heatmap}). These trends are naturally explained by decomposing the \FT{} $\rightarrow$ \UN{} difference at layer $l$ into a global and an example-specific component, $\Delta_i^{(l)} = h_{\text{UN}}^{(l)}(i)-h_{\text{FT}}^{(l)}(i) = g^{(l)} + \varepsilon_i^{(l)}$. The approximately constant centroid distance indicates that the norm of the global shift $|g^{(l)}|$ is stable across depth, while the gradual increase in pairwise distance reflects a growing example-specific term $\varepsilon_i^{(l)}$. Visualizing these dynamics via heatmaps reveals that while most languages follow this pattern, Chinese is a notable outlier, exhibiting a significant spike in centroid distance across intermediate layers (approx.\ layers 8--18) before realigning with the decoding-layer bottleneck. Furthermore, extending the analysis to the Base model (\textbf{FT vs Base} and \textbf{UN vs Base}) confirms that the unlearned model does not simply revert to the pre-trained state. Instead, the \textbf{UN vs Base} distance remains substantial, comparable to the \textbf{FT vs Base} shift, indicating that the unlearned model occupies a distinct representational state that retains general fine-tuning distributions while selectively suppressing specific knowledge via targeted output distortions.

\label{app:CPD}

\section{Steering Vector Algorithms}
In this section, we provide the pseudocode for the steering vector extraction and injection procedures described in Section~\ref{sec:SteeringVectorText}. 

Algorithm~\ref{alg:steering-vector} outlines the process of extracting the global unlearning direction $\mathbf{g}^{(l)}$. This is achieved by computing the difference between the normalized hidden states of the unlearned model (\UN) and the fine-tuned model (\FT), averaged over an auxiliary forget set. This procedure isolates the geometric shift induced by unlearning while preventing the leakage of specific answers that we intend to probe for recovery. Upon completion, this algorithm yields a vector that steers representations toward the unlearned behavior.

Algorithm~\ref{alg:steering-injection} details the inference-time intervention used to test reversibility. Here, the extracted steering vectors are \emph{subtracted} from the hidden states of the unlearned model during the forward pass. Consequently, this steers the model back toward the fine-tuned behavior by cancelling out the unlearned geometric shift. This effectively shifts the model's internal state back toward the fine-tuned distribution, allowing it to reproduce the forgotten knowledge.

\begin{algorithm}[h]
\caption{Extraction of Layer-wise Steering Vectors}
\label{alg:steering-vector}
\begin{algorithmic}[1]
\REQUIRE Models: Fine-tuned $M_{\FT}$ and Unlearned $M_{\UN}$
\REQUIRE Forget dataset $\mathcal{D}$
\ENSURE Steering vectors $\{\mathbf{g}^{(l)}\}_{l=1}^L$
\STATE \textbf{Function} $\textsc{Norm}(\mathbf{v}) = \mathbf{v} / \|\mathbf{v}\|_2$
\STATE Initialize accumulators $\mathbf{s}^{(l)} \gets \mathbf{0}$ for $l = 1,\dots,L$
\STATE Compute hidden states for $M_{\FT}$ and $M_{\UN}$ on dataset $\mathcal{D}$
\FOR{$l = 1$ to $L$}
    \FOR{each sample $x \in \mathcal{D}$}
        \STATE Let $t$ be the index of the last token in $x$
        \STATE $\mathbf{h}_{\FT} \gets \text{hidden state of } M_{\FT} \text{ at layer } l, \text{ token } t$
        \STATE $\mathbf{h}_{\UN} \gets \text{hidden state of } M_{\UN} \text{ at layer } l, \text{ token } t$
        
        \STATE $\tilde{\mathbf{h}}_{\FT} \gets \textsc{Norm}(\mathbf{h}_{\FT})$
        \STATE $\tilde{\mathbf{h}}_{\UN} \gets \textsc{Norm}(\mathbf{h}_{\UN})$
        
        \STATE $\mathbf{s}^{(l)} \gets \mathbf{s}^{(l)} + (\tilde{\mathbf{h}}_{\UN} - \tilde{\mathbf{h}}_{\FT})$
    \ENDFOR
\ENDFOR
\FOR{$l = 1$ to $L$}
    \STATE $\mathbf{g}^{(l)} \gets \textsc{Norm}(\mathbf{s}^{(l)})$ 
\ENDFOR
\STATE \textbf{return} $\{\mathbf{g}^{(l)}\}_{l=1}^L$
\end{algorithmic}
\end{algorithm}

\begin{algorithm}[t]
\caption{Layer-wise Steering Injection}
\label{alg:steering-injection}
\begin{algorithmic}[1]
\REQUIRE Unlearned model $M$ with layers $f_1,\dots,f_L$, steering vectors $\{\mathbf{g}^{(l)}\}$
\REQUIRE Hyperparameters: scale $\alpha > 0$, steering window size $N + 1$
\REQUIRE Evaluation batch $\mathcal{X} = \{(x_i, y_i)\}_{i=1}^B$
\ENSURE Optimal start layer $c^*$
\STATE $s_{\text{best}} \gets -\infty$
\STATE $c^* \gets \text{None}$
\FOR{$c = 1$ to $L - N$}
    \STATE $S_{\text{batch}} \gets 0$
    \FOR{$i = 1$ to $B$}
        \STATE $seq \gets x_i$ 
        \STATE $gen \gets \text{empty string}$
        \REPEAT
            \STATE $t \gets \text{length}(seq)$
            \STATE $H^{(0)} \gets \text{Embed}(seq)$
            \FOR{$l = 1$ to $L$}
                \STATE $H^{(l)} \gets f_l(H^{(l-1)})$
                \IF{$c \le l \le c + N$}
                    \STATE $\mathbf{h} \gets H^{(l)}[t]$
                    \STATE $r \gets \|\mathbf{h}\|_2$
                    \STATE $H^{(l)}[t] \gets \mathbf{h} - \alpha \cdot r \cdot \mathbf{g}^{(l)}$
                \ENDIF
            \ENDFOR
            \STATE $token \gets \textsc{Greedy}(H^{(L)}[t])$
            \STATE $seq \gets \text{concat}(seq, token)$
            \STATE $gen \gets \text{concat}(gen, token)$
        \UNTIL{$token$ is EOS}
        \STATE $S_{\text{batch}} \gets S_{\text{batch}} + \text{NLI}(y_i, gen)$
    \ENDFOR
    \STATE $\bar{s} \gets S_{\text{batch}} / B$
    \IF{$\bar{s} > s_{\text{best}}$}
        \STATE $s_{\text{best}} \gets \bar{s}$
        \STATE $c^* \gets c$
    \ENDIF
\ENDFOR
\STATE \textbf{return} $c^*$
\end{algorithmic}
\end{algorithm}

\section{Gemma Steering Vector Results}
\label{app:gemmaSteeringVectorResults}
\begin{figure*}[t]
  \centering
  \includegraphics[width=\linewidth]{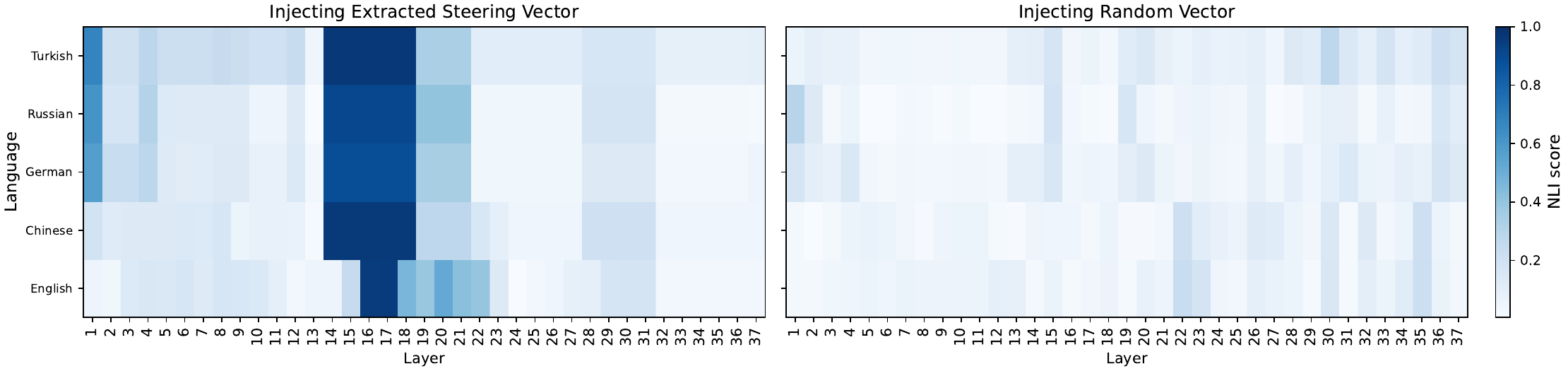}
  \caption{%
    Gemma2-9B steering injection results. Effect of layer-wise steering on NLI score for the forget set. Each heatmap shows the change in score when injecting a normalized steering direction with scale $\alpha$, where $\alpha$ is selected separately for each language. The left heatmap uses extracted steering vectors, which recover almost all of the forgotten knowledge when applied in middle and late layers. The right heatmap uses random Gaussian directions with matched norm and scale, the much weaker recovery indicates that the extracted vectors capture a structured unlearning direction rather than generic noise. This result is consistent with, yet more pronounced than, that of the Qwen2.5-7B model.
  }
  \label{fig:gemma_steer_result}
\end{figure*}

To evaluate the generalizability of our steering intervention across model families, we apply the inference-time recovery method to \emph{Gemma2-9B}. The experimental setup follows the procedure outlined in Section~\ref{sec:SteeringVectorText}, using steering vectors derived exclusively from an auxiliary English forget set. For these experiments, we apply the steering intervention over a window of $N=6$ consecutive transformer blocks to account for the model's greater depth compared to Qwen2.5-7B. For the hyperparameter $\alpha$, we perform a grid search and select $\alpha=0.8$ for English, $\alpha=1.0$ for Chinese, and $\alpha=1.2$ for German, Russian, and Turkish (baseline $\alpha=1.0$). The results, presented in Figure~\ref{fig:gemma_steer_result}, the steering intervention on Gemma-2 restores nearly the full performance capability of the original fine-tuned model across languages.

\section{Additional Steering Vector Experiments}
\label{app:supplementarySteeringVector}

\subsection{Chinese as the Source Language}
\label{app:chineseSourceSteering}

In the main text, we construct steering vectors using English questions as the source language. To test whether the recovery effect depends on this choice, we repeat the same experiment using Chinese questions to extract the steering vectors while keeping the evaluation protocol unchanged. The resulting recovery remains effective across all target languages. For Qwen2.5-7B with DPO unlearning, the best recovery scores are $68.21\%$ for English, $49.89\%$ for Chinese, $41.67\%$ for German, $38.21\%$ for Russian, and $56.15\%$ for Turkish. These results suggest that the steering vector is not tied to English-specific lexical information. Instead, they are consistent with our interpretation that computing activation differences on identical inputs largely cancels out input-specific semantic content and captures a more general unlearning-induced direction in the shared representation space.

\subsection{Additional Unlearning Methods}
\label{app:additionalUnlearningSteering}

To test whether steering-vector recovery is specific to DPO-based unlearning, we further evaluate the same procedure on models unlearned with gradient ascent (GA) and negative preference optimization (NPO). We conduct this experiment on Gemma2-9B and tune the intervention strength $\alpha$ and injection layer for each language.

For GA, the steering vector recovers forgotten information across all five languages, with best scores of $55.77\%$ for English, $80.60\%$ for Chinese, $41.58\%$ for German, $62.48\%$ for Russian, and $52.11\%$ for Turkish. For NPO, we observe the same qualitative pattern, with best scores of $68.95\%$ for English, $63.34\%$ for Chinese, $57.54\%$ for German, $65.99\%$ for Russian, and $51.49\%$ for Turkish. These results indicate that the recoverability of forgotten information is not unique to DPO, but also appears under other fine-tuning-based unlearning methods.

\subsection{Simplified All-Layer Steering Intervention}
\label{app:simplifiedAllLayerSteering}

In our initial steering experiments, we injected steering vectors over selected layer windows, which introduced an additional hyperparameter controlling the injection range. We found that this procedure can be simplified by applying the per-layer steering vectors across all layers with a substantially smaller intervention strength $\alpha$. This removes the need to tune the layer-window hyperparameter and reduces the cost of hyperparameter search. 
\end{document}